\documentclass[journal]{IEEEtran}
\ifCLASSINFOpdf
\else
\fi

\usepackage[square,sort&compress,comma,numbers]{natbib}

\usepackage{caption}
\usepackage{color,array}
\usepackage{graphicx}
\usepackage{csquotes}

\usepackage{amsmath,amssymb,amsfonts}
\usepackage{mathrsfs}

\usepackage[linesnumbered,ruled,vlined]{algorithm2e}

\SetCommentSty{mycommfont}

\SetKwInput{KwInput}{Input}                % Set the Input
\SetKwInput{KwOutput}{Output}

\usepackage{graphicx}
\usepackage{textcomp}
\usepackage{xcolor}
\usepackage[percent]{overpic}

\usepackage{xspace}

\usepackage{epsfig}
\usepackage{pifont}

\usepackage{booktabs}
\usepackage{multirow}

\usepackage{colortbl}    
    
\usepackage{soul}
\usepackage{pgfplots}
\usepackage{csquotes}
\usepackage{cancel}

\usepackage{soul}

\usepackage{color, colortbl}
\definecolor{Gray}{gray}{1}

\usepackage{hyperref}
\hypersetup{
    colorlinks = true,
    linkcolor=blue
}

\usepackage{pgfplots}

\usepackage{wrapfig}

\definecolor{lime}{HTML}{A6CE39}
\DeclareRobustCommand{\orcidicon}{%
	\begin{tikzpicture}
	\draw[lime, fill=lime] (0,0) 
	circle [radius=0.16] 
	node[white] {{\fontfamily{qag}\selectfont \tiny ID}};
	\draw[white, fill=white] (-0.0625,0.095) 
	circle [radius=0.007];
	\end{tikzpicture}
	\hspace{-3mm}
}

\foreach \x in {A, ..., Z}{%
	\expandafter\xdef\csname orcid\x\endcsname{\noexpand\href{https://orcid.org/\csname orcidauthor\x\endcsname}{\noexpand\orcidicon}}
}

\begin{document}
\title{A Generalized Zero-Shot Quantization of Deep Convolutional Neural Networks via Learned Weights Statistics}

\author{Prasen Kumar Sharma\orcidA,~\IEEEmembership{} Arun Abraham\orcidB, and Vikram Nelvoy Rajendiran\orcidC 

\IEEEcompsocitemizethanks{
        © 2021 IEEE.\@ Personal use of this material is permitted. Permission from IEEE must be
obtained for all other uses, in any current or future media, including
reprinting/republishing this material for advertising or promotional purposes, creating new
collective works, for resale or redistribution to servers or lists, or reuse of any copyrighted
component of this work in other works.}

\thanks{Prasen Kumar Sharma was an intern with Samsung R\&D Institute India- Bangalore during this work. Part of this work was
done when he was at Indian Institute of Technology Guwahati, India.} \thanks{P. K. Sharma is with TensorTour Inc. E-mail: prasenkrsharma@gmail.com. (\textit{Corresponding author})}. 
\thanks{Arun Abraham and Vikram Nelvoy Rajendiran are with Samsung R\&D Institute India- Bangalore. Emails: \{arun.abraham, vikram.nr\}@samsung.com.}% <-this % stops a space
\thanks{Manuscript received April 19, 2005; revised August 26, 2015.}}


\markboth{Journal of \LaTeX\ Class Files,~Vol.~14, No.~8, August~2015}%
{Shell \MakeLowercase{\textit{et al.}}: Bare Demo of IEEEtran.cls for IEEE Journals}



\maketitle

\begin{abstract}
Quantizing the floating-point weights and activations of deep convolutional neural networks to fixed-point representation yields reduced memory footprints and inference time. Recently, efforts have been afoot towards zero-shot quantization that does not require original unlabelled training samples of a given task. These best-published works heavily rely on the learned batch normalization (BN) parameters to infer the range of the activations for quantization. In particular, these methods are built upon either empirical estimation framework or the data distillation approach, for computing the range of the activations. However, the performance of such schemes severely degrades when presented with a network that does not accommodate BN layers. In this line of thought, we propose a \textit{generalized zero-shot quantization} (GZSQ) framework that neither requires original data nor relies on BN layer statistics. We have utilized the data distillation approach and leveraged only the pre-trained weights of the model to estimate enriched data for range calibration of the activations. To the best of our knowledge, this is the first work that utilizes the distribution of the pre-trained weights to assist the process of zero-shot quantization. The proposed scheme has significantly outperformed the existing zero-shot works, \textit{e.g.}, an improvement of $\sim$ 33\% in classification accuracy for MobileNetV2 and several other models that are w \& w/o BN layers, for a variety of tasks. We have also demonstrated the efficacy of the proposed work across multiple open-source quantization frameworks. Importantly, our work is the first attempt towards the post-training zero-shot quantization of futuristic unnormalized deep neural networks.
%An ablation study has been presented at the end to illustrate the effect of BN folding while performing the post-training quantization.
\end{abstract}


% \fi
\IEEEpeerreviewmaketitle

\section{Introduction}

\IEEEPARstart{D}{eep} CNNs are renowned for their tremendous capabilities to learn robust and complex features to produce remarkable results. Over the last decade, they have been widely adopted in various fields such as Computer Vision \cite{intro1}, Speech Processing \cite{intro2}, Natural Language Processing \cite{intro3}, \textit{etc.}, spanning from system to on-device platforms. However, several modern deep neural networks have an enormous size, which makes it challenging to deploy those on real-time resource constraint devices. For \textit{e.g.}, LeCun \textit{et al.} \cite{lenet5} proposed LeNet5 model that has approx. 400K weights. About two decades later, AlexNet \cite{alexnet} and VGG-16 \cite{vgg16} were proposed that have more than 60M, 130M weights, respectively. Each of these models, when stored in 32-bit floating-point (FP32) format, requires excessive storage, \textit{e.g.}, VGG16 \cite{vgg16}, that has a size of 130M$\times$4Bytes $\simeq$ 520MBytes. Coates \textit{et al}. \cite{coates} even proposed a model that has 11B weights. Such models may also result in higher inference latency, processing capacity, energy consumption, and difficulty in training \cite{rate-distortion}. 

Furthermore, a majority of the inference specific accelerators, such as Google Edge TPU, support only the quantized models. For \textit{e.g.}, MobileNetV2 \cite{mobilenetv2}, that takes 51 milliseconds (ms) per inference for ImageNet \cite{imagenet} classification on a Desktop CPU, takes  only 2.6 ms on a device with Google Edge TPU, about 20$\times$ faster. 
%\footnote{\url{https://coral.ai/docs/edgetpu/benchmarks/}} 
Therefore, it becomes necessary to compress the large-sized models in order to accommodate them on resource constraint devices and accelerators for an efficient real-time computing. 

Existing methods for deep model compression are primarily based on matrix factorization, pruning, quantization, and designing deep CNN's with fewer parameters \cite{pruning1,pruning2,pruning3,pruning4,OCS, nndesign1,nndesign2,nndesign3,nndesign4,mobilenetv2}.
 Also, quite a few schemes \cite{bayes1, bayes2, bayes3} have been proposed that train a deep CNN based on Bayesian methods in order to aid the quantization and pruning at a later stage. However, in this work, we focus on quantization and propose a novel principled zero-shot approach.
 
\textbf{Background:} Quantization allows us to store or process the tensors at lower and ultra-low bit precisions, \textit{e.g.}, $<$4, 8, 16$>$-bit fixed-point from FP32 format \cite{google-whitepaper}. It yields more resource-efficient integer operations. Given a tensor ${x}$ (in \textsc{FP32}\xspace), its quantized value ${Q}({x})$ can be approximated as
\begin{equation}
{Q}({x}) = \texttt{round} \big(x/\Delta + z\big),
\end{equation}
 where $\Delta$, $z$ denote scale and zero-point offset, respectively. The $\Delta$ can be computed as 
\begin{equation}
\Delta = n/(\texttt{max}({x}) - \texttt{min}({x})),
\label{eq:delta}
\end{equation} 
 where $n = 2^{b} - 1$ with $b$ denoting lower bit precision. The zero-point offset then can be computed as follows 
\begin{equation}
z = \Delta * \texttt{min}({x}).
\label{eq:z}
\end{equation}

Overall, the quantization schemes can be classified into (a) Per-tensor (QS-PT), and (b) Per-channel (QS-PC), alias Per-axis.  Per-tensor quantization requires $\Delta$ and $z$ to be computed across the tensor. Whereas in per-channel, they are calculated across each channel of the tensor. Further, they are categorized as symmetric if $z$ is 0, asymmetric (\textit{alias affine}) otherwise \cite{google-whitepaper}. 
%It should be noted that quantization schemes may also incur noise on weights and activations, which may severely degrade the performance.
We refer the readers to \cite{google-whitepaper, cvpr18-Efficient-Integer-Arithmetic-Only} for a detailed review of various quantization schemes. 

In the case of post-training quantization, $\Delta$ and $z$ are computed for both weights and activations prior to inference. Their computations (see Eqns.~\ref{eq:delta}, and \ref{eq:z}) depend on range of the input tensor ${x}$, \textit{i.e.}, $\texttt{min}({x})$, and $ \texttt{max}({x})$. We refer the process of estimating the range of the input tensor as ``\textit{range calibration}'' in the rest of the paper. Given a pre-trained model in FP32, range calibration of the weights can be achieved without any difficulty. However, for activations\footnote{interchangeably referred to as \textit{features} in the paper}, one requires access to the original unlabelled training samples. Once the dataset, either in full or limited, is available, the activations can be generated, followed by computing the $\Delta$ and $z$ of the same. Based on the availability of the original unlabelled training samples, either in full, limited, or none, the recent works can be categorized into (a) quantization-aware training \cite{google-whitepaper, cvpr18-Efficient-Integer-Arithmetic-Only, pact, xnor, lqnet, inq, dorefa, CAT}, (b) requires limited data \cite{NIPS19, ICCVW19, OCS, same}, and (c) data-free approaches \cite{dfq, zeroq,knowledge-within,inceptionism, gdfq}, respectively. 

\textbf{Challenges and Motivation:} Recently, much attention has been given to the data-free approaches \cite{dfq, zeroq,knowledge-within, gdfq} since the availability of the original data is infeasible in some cases. For \textit{e.g.}, Medical Imaging, where the user's privacy is prioritized above all. These schemes may also be analogously referred to as zero-shot approaches\footnote{since they do not require original data samples.}. For estimating the $\Delta$ and $z$ of the activations w/o utilizing the original data, these zero-shot schemes mainly rely on the learned BN \cite{bn} layer's parameters. The long-term statistics of BN \cite{bn} layer may consist of some information about the original training samples \cite{knowledge-within}, which may be useful for range calibration of the activations. 

BN \cite{bn} layer in a deep CNN is an indispensable component that is believed to improve the generalization, accelerate convergence,  aid higher learning rate, and stabilize the training process \cite{fixup}. However, in recent studies \cite{fixup, isonet, nfnets}, it has been shown that none of these benefits is unique to the BN \cite{bn} layer. Also, BN \cite{bn} parameters are folded into the weights and biases of convolution and fully-connected layers for an efficient inference, which is a widely adopted practice \cite{google-whitepaper}. The BN \cite{bn} folding introduces high variance into the weights of the pre-trained model, that makes quantization difficult \cite{google-whitepaper}. Hence, the existing zero-shot approaches \cite{dfq, zeroq} fail when the BN \cite{bn} layer is either folded or absent in the network. 

Now, an obvious question that arises, is, ``{what else can be utilized from the pre-trained model that may represent original training samples, given the zero-shot condition holds}.'' {Perhaps, the pre-trained weights}. For any given task, weights are learned to extract the imperative features from the input training samples in order to achieve the desired objective with minimal penalty \cite{visualizing}. 

Further, when the BN \cite{bn} layer is folded, the introduced variance in the weights may have some auxiliary information about the distribution of the original data. Both the pre-trained weights and folded-variance have not been exploited in any of the recent works \cite{google-whitepaper, cvpr18-Efficient-Integer-Arithmetic-Only, pact, xnor, lqnet, inq, dorefa, CAT, NIPS19, ICCVW19, OCS, same, dfq, zeroq}. 

Therefore, in this work, we have proposed a generalized zero-shot approach, namely GZSQ, that neither requires original training samples nor depends on BN \cite{bn} layer parameters \textcolor{black}{for range calibration of the activations}. The proposed GZSQ acts as an API call and is built upon the data distillation framework that only considers the pre-trained weights of the model. \textcolor{black}{In particular, GZSQ approximates the substitutes for BN statistics for each layer by only utilizing the pre-trained weights of the network. 
Hence, it eliminates the requirement of the BN layers in the model if one wants to incorporate data distillation for the range calibration of the activations under zero-shot condition.} We have evaluated the proposed method across multiple quantization frameworks with both QS-PC and QS-PT schemes. Further, in our experiments, we have considered the models that are w and w/o BN \cite{bn} layers, including the BN \cite{bn} folding in the former case, for the tasks of classification \cite{imagenet, cifar10} and object detection \cite{coco}. 

The rest of the paper is organized as follows: Section \ref{related_works} briefly reviews the related existing works. Section \ref{gzsq} presents the proposed GZSQ framework. The experimental details, results, and ablation study are presented in Sections \ref{experiments}, \ref{results}, and \ref{ablation}, respectively. Lastly, Section \ref{conclusion} summarizes the work.
 
\section{Related Developments}
\label{related_works}
Besides quantization, several works have been proposed to address the restricted memory footprints and inference latency of the modern deep CNNs. Going orthogonally to quantization, these methods are based on $-$ efficient neural architecture design, knowledge distillation, pruning, hardware co-design, and matrix factorization.
 
\textit{\textbf{Architecture design-based}} methods  \cite{nndesign1,nndesign2,nndesign3,nndesign4,mobilenetv2}    
 proposed the deep CNNs that have fewer number of parameters using Neural Architecture Search (NAS) \cite{nas}, for \textit{e.g.}, MobileNetV2 \cite{mobilenetv2}. \textit{\textbf{Knowledge distillation based}} methods \cite{fitnets,books, snidhi} extracted the latent information from the original model using student-teacher paradigm and proposed the sparse version of the model. To reduce the number of trainable parameters, \textit{\textbf{pruning}} \cite{pruning1,pruning2,pruning3,pruning4} sets the insignificant weights to zero. This includes methods based on Hessian matrix of the loss functions \cite{obd, obs}, and iterative pruning \cite{iterative}. Kwon \textit{et al}. \cite{co-design} proposed to \textbf{\textit{co-design}} the deep CNNs and neural net accelerators \cite{nna} to achieve significant performance gain. \textit{\textbf{Matrix factorization based}} schemes \cite{denton, cheng} decompose the weight matrix into the low-rank ($\mathbf{L}$) and sparse ($\mathbf{S}$) matrices, where both $\mathbf{L}$ and $\mathbf{S}$ require less storage. Later, Han \textit{et al}. \cite{han_combine} proposed a method based on the ensemble of pruning, quantization and Huffman coding \cite{huffman}. Recently, He \textit{et al.} \cite{rl} proposed the reinforcement learning-based method for model compression. However, here, we focus on existing quantization works based on the requirement and availability of the original training samples. 

\textbf{\text{Require full training set}:} Quantization reduces the memory footprints of the deep CNNs by reducing the bit precisions of weights and activations. However, it may also introduce noise and lead to the significant performance degradation, especially when quantizing FP32 to ultra-low bit precisions. To address this, several methods \cite{google-whitepaper, tmm1, tmm2, cvpr18-Efficient-Integer-Arithmetic-Only, pact, xnor, lqnet, inq, dorefa, CAT} follow the trivial approach for quantization by performing the Quantization-aware Training (QAT) \cite{cvpr18-Efficient-Integer-Arithmetic-Only}. These approaches may result in the lowest degradation in final accuracy. However, it requires a full training set to train the model, which may not be possible in some cases, \textit{e.g.}, Medicine or Satellite Image-based systems, due to privacy. Besides the availability of original data, QAT-based schemes are time and resource consuming.

\textbf{\text{Require limited data}:} To overcome this drawback, a few methods \cite{NIPS19, ICCVW19, OCS, same}, based on weights factorization and channel splitting, have been proposed, requiring {limited training samples}. In particular, Choukroun \textit{et al}. \cite{ICCVW19} proposed a method, namely OMSE, that minimizes the $\ell_{2}$ distance between quantized and original tensor. Also, Banner \textit{et al}. \cite{NIPS19} proposed a method called ACIQ that analytically computes the clipping range and per-channel bit allocations for deep CNNs. However, per-channel quantization for activations is infeasible in practice. Zhao \textit{et al}. \cite{OCS} proposed a method called OCS to solve the problem of outlier channel.

\textbf{Zero-shot approaches:} To overcome the drawback of original data dependency, either in full or limited, Nagel \textit{et al}. \cite{dfq} proposed a data-free quantization method, namely \textsc{DFQ}. It is based on the weights equalization and bias correction. However, to estimate the $\Delta$ and $z$ of the activations, \textsc{DFQ}\xspace entirely depends on the \textsc{BN}\xspace 's \cite{bn} shift ($\beta$) and scale ($\gamma$) parameters. The $j^{th}$ layer activation ranges are first calculated as $\beta_j \pm n.\gamma_j$, where $n$ has been empirically set to 6. Later, $\Delta_j$ and $z_j$ have been computed using Eqns.~\ref{eq:delta} and \ref{eq:z}, respectively. Subsequently, following \cite{knowledge-within,inceptionism}, Cai \textit{et al}. \cite{zeroq} proposed a zero-shot framework, called \textsc{ZeroQ}\xspace, to generate the data that matches the statistics of the training set. For this, \textsc{ZeroQ} \cite{zeroq} has utilized the BN \cite{bn} layer's statistics, as
\begin{equation}
    \min_{y^r} \sum_{i=0}^L \|\tilde \mu_i^r - \mu_i\|_2^2 + \|\tilde \sigma_i^r - \sigma_i\|_2^2,
\label{eq:synthetic_gaussian}
\end{equation}
where $\mu_i^r$/$\sigma_i^r$ are the mean/std-dev of the 
distilled data's ($y^r$) distribution at layer $i$, and $\mu_i$/$\sigma_i$ are the corresponding mean/std-dev parameters stored in the $i^{th}$ BN \cite{bn} layer. The distilled data $y^r$ then has been used to infer the range of the activations to estimate the $\Delta$ and $z$. Later, Xu \textit{et al.} \cite{gdfq} proposed a generative approach \cite{gan} to produce a fake data for range calibration by exploiting the classification boundary knowledge and BN \cite{bn} statistics. The proposed generator architecture is inspired from AC-GAN \cite{acgan}, which also results in additional memory overhead. 

Recently, a shift in trend has been observed towards the Mixed-Precision Quantization (MPQ) \cite{mp1,mp2,mp3}, where a different bit-precision is set adaptive to each layer in the network. However, it further raises another level of difficulty in finding the optimal bit-width for each layer, given a predetermined support by the hardware accelerators to MPQ. Along with DFQ \cite{dfq}, {these best-published works fail when the model does not comprise of BN \cite{bn} layers. In that case, it becomes difficult to either utilize the distillation approach or BN \cite{bn} parameters ($\beta, \gamma$) and infer activation ranges without  original training samples.} 

Moreover, the trend \cite{fixup,isonet, nfnets} observed in the past two years shows that the community has started to look for alternatives to BN \cite{bn}. Therefore, it should be mentioned that in the coming years, a majority of the cutting-edge deep CNNs may not consist of BN \cite{bn} layers at all. This work is the first attempt towards the quantization of such futuristic models.

\textbf{Our Contributions:}
 We present a novel solution that allows us to utilize the distillation approach, even if the BN \cite{bn} layers are folded or absent in a model. {For this, we first estimate the substitutes to BN statistics ($\mu, \sigma$) for every layer by utilizing the pre-trained weights of the model only}. {We divide this substitutes estimation phase into (a) statistics estimation, followed by (b) empirical statistics adjustment sub-phases. We then perform the data distillation, similar to in} Eqn.~\ref{eq:synthetic_gaussian}, {by utilizing the estimated substitutes with a novel objective function}. Essentially, our key contributions are fourfold: 

\begin{itemize}

\item We propose a \textit{generalized zero-shot quantization} (GZSQ) framework that neither requires original data nor relies on BN \cite{bn} layers statistics of the models. In other words, we leverage the pre-trained weights of the model ``only'' and estimate an enriched data which can be used for range calibration of the activations.

\item  We also propose to use the absolute Z-score based loss function over $\ell_1$ and $\ell_2$ norms or Kullback-Leibler (KL) divergence for an efficient data distillation.

\item To test the generalization proficiency of the proposed GZSQ, we have evaluated our distillation approach across multiple open-source quantization frameworks. We have also benchmarked the models that are w \& w/o BN \cite{bn} layers on the tasks of classification and object detection. 
%A glimpse of the obtained results is shown in Fig.~\ref{intro-fig}. 

\item To appraise the efficacy of the proposed method, we have shown the results on a different problem domain, \textit{e.g.}, Medical Imaging. Moreover, we have presented an ablation study to demonstrate the effect of various cost functions and BN \cite{bn} parameters folding before \& after data distillation.

\end{itemize}

\section{Proposed Approach}
\label{gzsq}

\begin{table}[h]
\centering
\captionsetup{textfont={sc,footnotesize}, justification=centering, labelsep=newline}
\caption{Notations}
\begin{tabular}{!{\color{black}\vrule}c!{\color{black}\vrule}l!{\color{black}\vrule}}
\arrayrulecolor{black}\hline
\textcolor{black}{{Symbol}} & \textcolor{black}{{Meaning}}\\
\hline
\textcolor{black}{$T$} & \textcolor{black}{A tensor that  $\in \mathbb{R}^{c\times h \times w}$}\\ 
\hline
\textcolor{black}{${\mu}_{T}$, ${\sigma}_{T}$} & \textcolor{black}{Channel-wise mean and std-dev of ${T}$, respectively} \\
\hline
\textcolor{black}{${F}$} & \textcolor{black}{A pre-trained FP32 model with $N$ conv2d layers}\\
\hline
\textcolor{black}{$\mathbf{W}$} & \textcolor{black}{Pre-trained conv2d weights (in FP32) of ${F}$}\\ 
\hline
\textcolor{black}{$\mathbf{W}_n$} & \textcolor{black}{Weights of $n^{th}$ conv2d layer in $F$}\\
\hline
\textcolor{black}{$f^n$} & \textcolor{black}{$n^{th}$ layer activations in ${F}$} \\
\hline
\textcolor{black}{$\mu_{f^n}$, $\sigma_{f^n}$} & \textcolor{black}{Mean and std-dev of $f^n$}\\
\hline
\textcolor{black}{$\mu_{a^{1:N}}$, $\sigma_{a^{1:N}}$} & \textcolor{black}{Substitutes of BN \cite{bn} statistics}\\
\hline
\textcolor{black}{$\mathbb{A}$} & \textcolor{black}{Set consisting of BN \cite{bn} substitutes}\\
\hline
\textcolor{black}{$\mathcal{N}(\mu, \sigma)$} & \textcolor{black}{Gaussian with $\mu$ mean and $\sigma$ std-dev}\\
\hline
\end{tabular}
\label{tab:notations}
\end{table}
In this section, \textcolor{black}{following the notations in Table \ref{tab:notations}}, we describe the regime of operations of the proposed GZSQ scheme.
%\textit{\textbf{Notations:}} Let for a given tensor ${T}\in \mathbb{R}^{c\times h \times w}$, ${\mu}_{T}$ and ${\sigma}_{{T}}$ denote its channel-wise mean and std-dev, respectively.  We denote the pre-trained {conv2d} weights (in FP32) of a model ${F}$, with $N$ such layers, as $\mathbf{W}$. The weights of ${n}^{th}$ {conv2d}  layer in ${F}$ can be denoted as $\mathbf{W}_n$. We also denote the corresponding $n^{th}$ layer activations as $f^{n}$. The mean and std-dev of $f^{n}$ then can be denoted as $\mu_{f^{n}}$ and $\sigma_{f^{n}}$. We denote the substitutes of BN \cite{bn} statistics as $\mu_{a^{{1:N}}}$ and $\sigma_{a^{{1:N}}}$, and refer as $\mathbb{A}$.
 
\subsection{Statistics Estimation}
\label{sec:se}
We first describe the process of estimating $\mathbb{A}$, for ${F}$ w/o using original data samples. Following \cite{tradi}, if $\mathbf{W}_{1:N}$ are initialized with a Gaussian distribution ($\mathcal{N}(\mu,\sigma)$) \cite{xavier} \footnote{In practice, weights of many deep CNNs are initialized with Gaussian version of the Kaiming  initialization \cite{kaiming}.}, owing to the Central Limit Theorem, $\mathbf{W}_{1:N}$ will also converge to $\mathcal{N}(\mu,\sigma)$ at the end of training. We also assume the input $\mathbf{J}$ for distilled data estimation to follow $\mathcal{N}(0,1)$. 

Now, let us first consider the base case where in ${F}$, BN \cite{bn} layers exist upto layer $n$, where $n \in (1, N)$. The activations of layer $n$ then can be written as
\begin{equation}
f^{n} = {g}^n\big(\text{BN}^n\big[\mathbf{W}_n * f^{n-1} + \mathbf{b}^n\big]\big),
\end{equation}
where ${g}^n$ denotes $n^{th}$ layer activation function. Knowing that the  pre-activations of layer $n$ follow $\mathcal{N}(\mu,\sigma)$ due to $\text{BN}^n$ \cite{dfq}, and  if ${g}^n$ is some form of the class of clipped linear activations, such as ReLU or ReLU6, then $f^{n}$ will also follow a clipped ($i,j$) Normal distribution, where $i < j$, and $j$ can be $\infty$. 

Now, let us assume, from ${n+1}^{st}$ layer onwards, BN \cite{bn} layers do not exist in ${F}$. The activations of ${n+1}^{st}$ layer can be written as
\begin{equation}
f^{n+1} = {g}^{n+1}\big[\mathbf{W}_{n+1} * f^{n} + \mathbf{b}^{n+1}\big].
\end{equation}
\noindent We know that $\mathbf{W}_{n+1}$ and $f^{n}$ follow $\mathcal{N}(\mu,\sigma)$. Following the property of convolution of two Gaussian distributions $\mathcal{N}(p_1,q_1)$, $\mathcal{N}(p_2,q_2)$ 
is also a Gaussian $\mathcal{N}(p_1+p_2,\sqrt{q_1^2 + q_2^2})$\footnote{The readers are advised to refer this \href{http://citeseerx.ist.psu.edu/viewdoc/download?doi=10.1.1.583.3007\&rep=rep1\&type=pdf}{link} for more details.}, the $f^{n+1}$ will also follow a clipped $\mathcal{N}(\mu,\sigma)$ when $\mathbf{b}^{1:N}$ = 0 \footnote{The pre-trained models we have studied in this paper do not use bias.}. However, we will consider the extra biases introduced due to BN \cite{bn} folding in Section \ref{sec:bnf}. 

The substitutes  $\mu_{a^{n+1}}$ and $\sigma_{a^{n+1}}$ then can be estimated as follows
\begin{equation}
\label{mean_std_formula}
\begin{split}
\mu_{a^{n+1}} &= \mu_{\mathbf{W}_{n+1}} + \mu_{a^{n}}\\
\sigma_{a^{n+1}}&=\sqrt{\sigma^{2}_{\mathbf{W}_{n+1}} + \sigma^{2}_{a^{n}}}.
\end{split}
\end{equation}
\noindent This can go on for the subsequent layers ($n+2:N$) in ${F}$. Backtracking our initial assumption of BN \cite{bn} layers upto $n^{th}$ layer in ${F}$ to the $1^{st}$ layer, the $\mu_{a^{2}}$, $\sigma_{a^{2}}$ can be estimated following the above procedure. 

Now, suppose, even the $1^{st}$ layer in ${F}$ is not followed by a BN \cite{bn} layer. Then its activations can be written as $f^{1} = {g}^1\big[ \mathbf{W}_1 * \mathbf{J} \big]$. Holding to our initial assumption of $\mathbf{J}$ to follow $\mathcal{N}(0,1)$, the $f^{1}$ will also follow $\mathcal{N}(\mu,\sigma)$ and $\mu_{a^{1}}$, $\sigma_{a^{1}}$ can be estimated using Eqn.~\ref{mean_std_formula}.

\subsection{Empirical Statistics Adjustment}
\label{sec:esa}
The channel-wise addition in Eqn.~\ref{mean_std_formula} is true, iff the number of channels (\text{C}) of the addends are same. However, with the recent advancement in deep learning, there exists a variety of networks where the condition holds false \cite{mobilenetv2, resnet, shufflenet,inceptionv3}. We classify such pair of layers into the sets of (a) \text{Expansion} when \text{C}($\mathbf{W}_{n+1}$)    $>$ \text{C}($a^n$), and (b) \text{Contraction} when \text{C}($\mathbf{W}_{n+1}$)    $<$ \text{C}($a^n$). We empirically define the following set of values as \{\text{min, mean $\pm$ min, mean, max $\pm$ mean, max}\}, to be used for accommodating the required number of channels in the addend with lower \text{C}. 

Intuitively, a pair of filters of a deep CNN layer may have some correlation (\text{strong} or \text{weak}), resulting in correlated feature maps \cite{ps_wacv}. Therefore, the best approximation of the required statistics from the current ones w/o utilizing any explicit information may be made by leveraging any value from the set mentioned above. 

For \textit{e.g.}, in the case of \text{expansion}, repeating the current statistics across the channels works for most of the networks, except for \text{ShuffleNet} \cite{shufflenet} and \text{InceptionV3} \cite{inceptionv3}-based models, where \text{mean}($\mu_{a^{n}}$; $\sigma_{a^{n}}$)$-$\text{min}($\mu_{a^{n}}$; $\sigma_{a^{n}}$) works. 

Similarly, for \text{contraction}, where in most cases \text{mean}($\mu_{a^{n}}$;$\sigma_{a^{n}}$)$-$\text{min}($\mu_{a^{n}}$;$\sigma_{a^{n}}$) works, except for \text{SqueezeNet} \cite{squeezenet} where mean values are set with \text{mean}($\mu_{a^{n}}$) and for \text{MobileNetV2} \cite{mobilenetv2} where std-dev values are set using \text{mean}($\sigma_{a^{n}}$)$+$\text{min}($\sigma_{a^{n}}$).

\subsection{Data Distillation}
\label{sec:dd}
With the set $\mathbb{A}$ as substitutes to BN \cite{bn} statistics, we have followed the distillation approach similar to \textsc{ZeroQ}\xspace \cite{zeroq}. For this, we input a random instance of $\mathcal{N}(0,1)$, denoted by $y$, to ${F}$ and generate set of activations for each layer. We then compute the statistics of each activation as ($\mu_{f^{n}}, \sigma_{f^{n}}$) $\forall n \in 1:N$, and minimize the distributional difference with the corresponding values in $\mathbb{A}$. However, instead of utilizing $\ell_1$ or $\ell_{2}$ norm, or KL divergence, we have considered the absolute difference of Z-score test as a loss function $\mathcal{L}_\mathbf{Z}(,.,)$ for an efficient knowledge distillation as
\begin{equation}
\mathcal{L}_\mathbf{Z} ({u}, {v}) = \|\mu_{u} - \mu_{v}\|/\sqrt{(\sigma_{u}+s)^2 + (\sigma_{v}+s)^2},
\label{kd-eq}
\end{equation}
\noindent where ${u}$, ${v}$ can be any two tensors with different distributions and $s$ = 1e-6\footnote{\textcolor{black}{Similar to \textsc{ZeroQ} \cite{zeroq}, it has been set to avoid the division by 0 in Eqn. \ref{kd-eq} if $\sigma_{u}$ and $\sigma_{v}$ are 0.}}. The $\ell_1$ and $\ell_2$ norms are known to favour the sparse and non-sparse data, respectively. It has been widely accepted that in a deep CNN, the first few activations may have high non-sparsity whereas later layers may have high sparsity \cite{sparse}.  The shortcomings of utilizing KL divergence in GZSQ have been discussed in Section \ref{cost_fn_ablation}.

Therefore, instead of data or sparsity based optimization, we propose a distributional minimization ($\mathcal{L}_\mathbf{Z}$) and define the final loss as 
\begin{equation}
\mathcal{L}_\mathbf{D} = \Big[ \sum_{n=1}^{N} \mathcal{L}_{\mathbf{Z}} (  f^n  ,   a^n  )\Big] + \mathcal{L}_\mathbf{Z} (y, \mathcal{N}(0,1)).
\label{eq:ld}
\end{equation}

We finetune $y$ using the following objective function
\begin{equation}
\arg\min_{\hat{y}} \mathcal{L}_{\mathbf{D}},
\end{equation}
and then use the $\hat{y}$ (distilled data w/o utilizing BN \cite{bn} statistics) to infer the ranges of the activations in ${F}$. 

\subsection{The Curious Case of Batch Normalization Folding}
\label{sec:bnf}
In practice, for a resource-efficient inference, the BN \cite{bn} parameters are folded into the weights and biases of the convolutional and fully-connected layers of a deep CNN. For \textit{e.g.}, folded weights and biases of the $n^{th}$ convolutional layer in ${F}$  can be written as 
\begin{equation}
\begin{split}
\mathbf{W}_{n,\texttt{fold}} &= \gamma_n. \mathbf{W}_n/\sigma^B_n \\ 
\mathbf{b}^n_{\texttt{fold}} &= \beta_n - \gamma_n.\mu^B_n/\sigma^B_n,
\end{split} 
\end{equation}
where $\gamma_n, \beta_n, \mu^B_n, \sigma^B_n$ represent the scale, shift, long-term mean, and long-term std-dev of the $n^{th}$ BN \cite{bn} layer, respectively. It has been observed during our experiments that the introduced bias, when BN \cite{bn} layers are folded before distillation, may lead to a slight (severe in some cases; \textcolor{black}{\textit{e.g.}, ResNet18 and ResNet50; see Section \ref{sec:effect_of_bnf} and Table. \ref{tab:bias_vs_no_bias}}) degradation in the accuracy of the quantized models. Therefore, the introduced bias term has to be accommodated into the proposed distillation approach (see Sections~\ref{sec:se}, \ref{sec:esa}), when folding is performed before distillation. With the $\sigma_{\mathbf{b}^n_{\texttt{fold}}}$ = 0, the refined mean in Eqn.~\ref{mean_std_formula} can be written as 
\begin{equation}
\mu_{a^{n}} = \mu_{\mathbf{W}_{n,\texttt{fold}}} + \mu_{a^{n-1}} + \mathbf{b}^n_{\texttt{fold}}.
\end{equation}

\begin{algorithm}[t]
\DontPrintSemicolon
  \KwInput{${F}$, in FP32}
  \KwOutput{$\hat{y}$}
  \KwData{\textcolor{black}{\texttt{None}}}
%  \tcc{Initialize a random seed}
\tcc{Details of the methods SE, ESA and DD are given in Subsections \ref{sec:se}, \ref{sec:esa}, and \ref{sec:dd}, respectively.}  
  $\mathbf{J} \leftarrow \mathcal{N}(0, 1)$
  
  Define set $\mathbb{A}$ as \{($\mu_{a^1}$, $\sigma_{a^1}$), ($\mu_{a^2}$, $\sigma_{a^2}$), ..., ($\mu_{a^N}$, $\sigma_{a^N}$)\} 
    
  $\mu_{a^1}$, $\sigma_{a^1} \leftarrow$ \textit{SE}(0, 1, $\mathbf{W}_1$)
  
  \While{$n\in 2:N$}
   {
   		$\mu_{a^{n-1}}$, $\sigma_{a^{n-1}} \leftarrow$ \textit{ESA}($\mu_{a^{n-1}}$, $\sigma_{a^{n-1}}$, $\mathbf{W}_n$)\;
   		$\mu_{a^n}$, $\sigma_{a^n} \leftarrow$ \textit{SE}($\mu_{a^{n-1}}$, $\sigma_{a^{n-1}}$, $\mathbf{W}_n$)\;
   		
   }
       \tcc{Initialize $y$ with a random instance}
  $y \leftarrow \mathcal{N}(0, 1)$\;
  \tcc{Update $y$}
  $\hat{y} \leftarrow$ \textit{DD}($y$, $\mathbb{A}$, ${F}$)\;
    
%    \tcc{Quantize the pre-trained weights of the model ${F}$}
%  \tcc{Utilize $\hat{\mathbf{R}}$ to generate the activations, record \texttt{min-max}, obtain $\Delta, z$, for each layer. Finally, use the obtained $\Delta, z$ for the quantization of the activations at inference for each test sample.}
\caption{GZSQ algorithm}
\label{algo1}
\end{algorithm}

\subsection{Post-Training Quantization}
By utilizing the distilled data $\hat{y}$ to infer the ranges of the activations, the scale $\Delta$ and zero-point offset $z$ can be computed for each layer activation. For $\mathbf{W}$, one can compute the ranges and $\Delta, z$ as usual. In this paper, we have presented the results for both QS-PC and QS-PT schemes. Also, to test the generalization capacity of the proposed GZSQ scheme, we have incorporated both (a) \textsc{ZeroQ}\xspace 's \cite{zeroq} quantization simulation\footnote{We use the quant. simulation of \textsc{ZeroQ}\xspace + distillation process of GZSQ.}, and (b) Pytorch's Post-Training Static Quantization\footnote{\textcolor{black}{\url{https://tutorials.pytorch.kr/advanced/static_quantization_tutorial.html}}}, in our experimentation. In the latter case, we have adopted the QS-PT (affine) scheme with histogram observer, considering the hardware practicality, for the activations. Whereas, for the weights, we have shown the results for both QS-PC (symmetric) with per-channel min-max observer \& QS-PT (affine) with min-max observer, schemes. Observers record the range of the tensor and use it to compute the quantization parameters ($\Delta$, $z$). The details are beyond the scope of this work, hence, we refer the readers to \footnote{\url{https://pytorch.org/docs/stable/torch.quantization.html}}.

An overview of the proposed GZSQ framework has been given in the Algorithm~\ref{algo1}.

\section{Experiments}
\label{experiments}
This section first describes the details of the datasets, covering the vast diversity across the tasks, that have been used to evaluate the quantized models. It then shifts the attention towards the details of experimental settings and the state-of-the-art baselines, which have been appraised against the proposed GZSQ framework.

\textit{\textbf{Datasets:}} 
We start our results (see: Section~\ref{results}) by showing the classification accuracy on CIFAR-10 \cite{cifar10} and ImageNet \cite{imagenet} datasets. CIFAR-10 consists of $\sim$ 60K images across 10 different classes, of which pre-determined 10K samples have been used for testing the accuracy of the quantized model.
 ImageNet is one of the largest available benchmark datasets for image classification that consists of $\sim$ 1.2M natural images for training and 50K for validation, across 1K classes. We have also presented the results for the task of object detection using Microsoft COCO \cite{coco} dataset. It consists of more than 200K images across 80 different object categories. 
 
\begin{wrapfigure}{L}{0.2\textwidth}
\centering
\setlength{\tabcolsep}{1pt}
\begin{tabular}{c}

\includegraphics[width=0.19\textwidth, height=2.5cm]{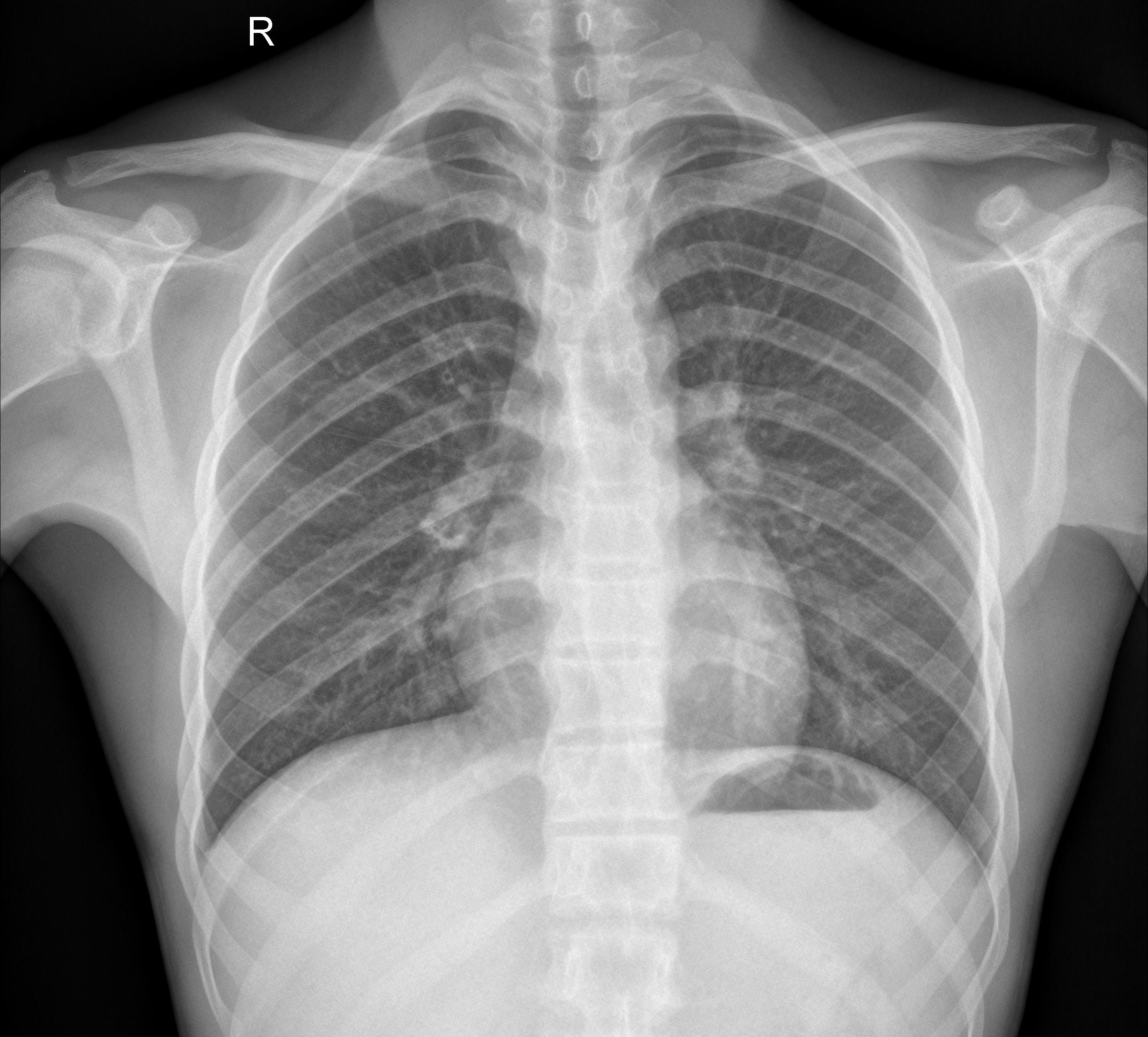}
\\

{\footnotesize \textbf{\textcolor{green}{Normal}}} \\

\includegraphics[width=0.19\textwidth, height=2.5cm]{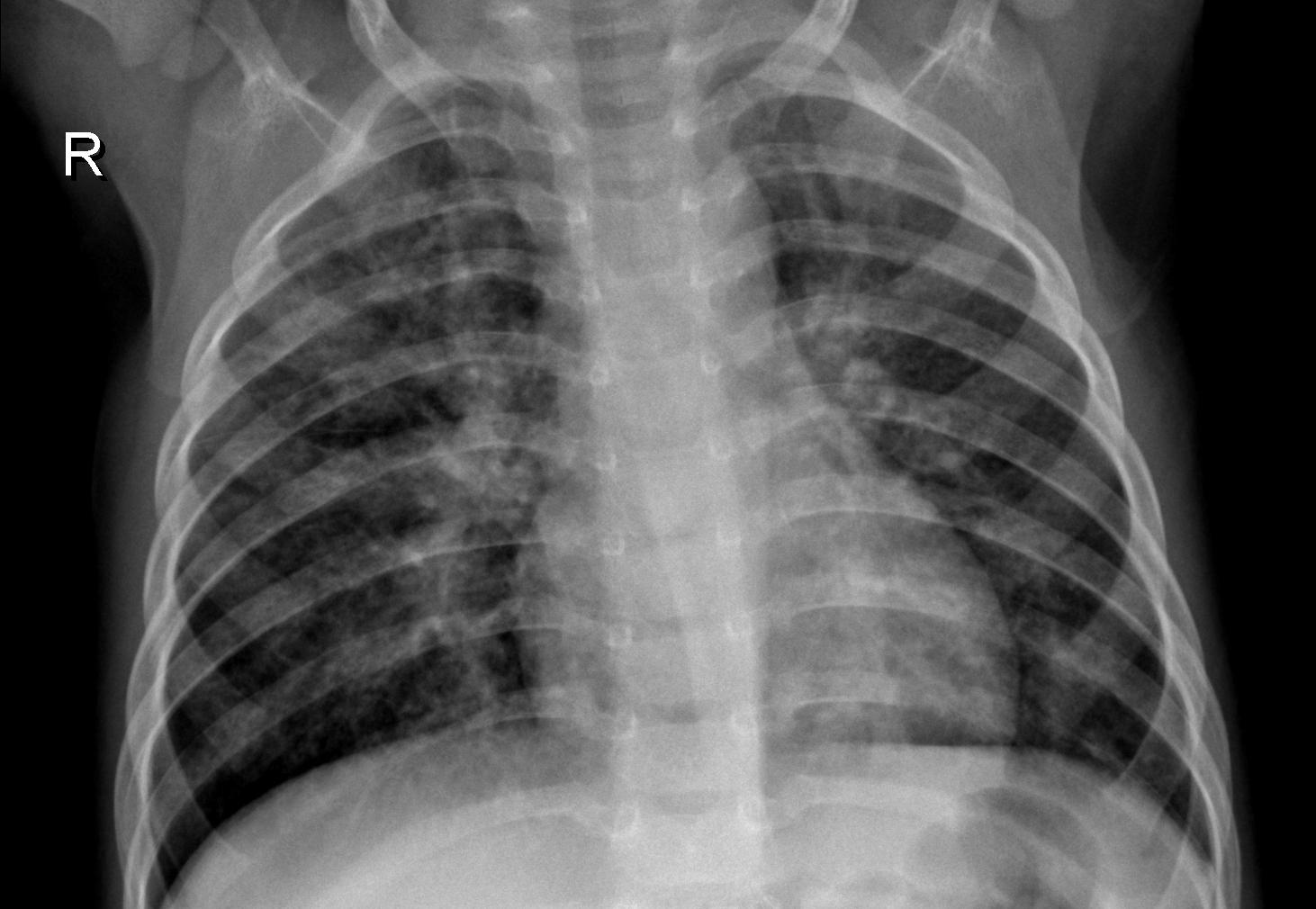}\\

 {\footnotesize \textbf{\textcolor{red}{Pneumonic}}}\\

\end{tabular}
\caption{Sample still of normal and pneumonic person's chest X-rays.}
\label{medical_samples}
\end{wrapfigure}

 Lastly, to test the efficacy of the proposed GZSQ, we have evaluated the quantized models in the domain of medical imaging for the task of pneumonia classification in chest X-rays. \textcolor{black}{The incorporated binary classification aims to categorize the given X-ray samples into normal and pneumonic classes. The pneumonic samples have been collected by utilizing both bacterial and viral infected cases. For this, we exploited the pre-split publicly available dataset\footnote{\url{https://www.kaggle.com/paultimothymooney/chest-xray-pneumonia}} \cite{pneumonia} consisting of $\sim$ 6K chest X-rays. The complete dataset has been divided into train, validation, and test subsets. The training subset consists of 5.2K X-rays with 1341 and 3875 samples corresponding to normal and pneumonic cases. The validation subset comprises 16 equally distributed radiographs. Whereas the test set contains 624 images with 234 and 390 samples corresponding to normal and pneumonic cases. Samples of chest X-rays of normal and pneumonic persons are shown in Figure~\ref{medical_samples}.} \textcolor{black}{The normal chest X-ray (see Figure~\ref{medical_samples}; \textit{top}) exhibits clear lungs without any abnormal opacified regions. Whereas the pneumonic X-ray (see Figure~\ref{medical_samples}; \textit{bottom}) typically depicts a focal lobar consolidation or interstitial pattern in one or both lungs.}

It should be mentioned that for most of the tasks, we first adopt the publicly available pre-trained models, perform GZSQ, and test the quantized models on the validation or test set. For the tasks where it is difficult to get the pre-trained models, we first train the models using the suggested configuration in the respective papers. Later, we perform GZSQ, and test the quantized models on the validation or test set. The proposed GZSQ scheme does not utilize either of the train, test, or validation samples in any form for any purpose at any stage to quantize a model.

%\begin{figure}[t]
%\end{figure}

\textit{\textbf{Experimental Environment:}} 
We have performed our experiments using Pytorch \cite{pytorch} framework with Adam \cite{adam} optimizer and a learning-rate of ${10^{-4}}$ for {500} iterations on Nvidia Tesla P100 GPU. In all our experiments, mean results have been presented with the corresponding std-devs across 10 runs, each with a different random seed. 

\textit{\textbf{Competing methods:}} 
We have evaluated the proposed scheme on various models that are with BN \cite{bn} layers, such as ResNets \cite{resnet}, MobileNets \cite{mobilenetv2}, ShuffleNets \cite{shufflenet}, InceptionV3 \cite{inceptionv3}, SqueezeNext \cite{nndesign1} for the task of image classification. We have also considered the models that are w/o BN \cite{bn} layers, such as Fixup-ResNets \cite{fixup} and ISONets \cite{isonet}, for image classification. RetinaNet \cite{retina} has been preferred for the task of object detection. For the case of medical imaging, we have considered the problem of Pneumonia classification in chest X-rays \cite{pneumonia}. For this, we have adopted ResNet18 \cite{resnet} and ResNext \cite{resnext} models. 

We have shown results against the recent best-published quantization works: RVQuant \cite{rvquant} (ECCV'18), DFQ \cite{dfq} (ICCV'19), OCS \cite{OCS} (ICML'19), ACIQ \cite{NIPS19} (NIPS'19), GDFQ \cite{gdfq} (ECCV'20), DFC \cite{knowledge-within} (CVPR'20), and the most recent zero-shot work, namely, \textsc{ZeroQ} \cite{zeroq} (CVPR'20). For a fair comparison, all methods were evaluated using respective default parameters.

\begin{table*}[t]
\captionsetup{textfont={sc,footnotesize}, justification=centering, labelsep=newline}
  \centering
      \caption{Top-1 accuracy on Fixup-ResNet \cite{fixup} models for the task of CIFAR-10 \cite{cifar10} classification. Quantization configuration is \text{W}p\text{A}q denoting the p-bit and q-bit precisions used for weights(QS-PC) and activations(QS-PT), respectively.  }
%  \caption{Add caption}
\resizebox{\textwidth}{!}{
    \begin{tabular}{|l|c|c|c|c|c|c|c|c|c|c|}
%   \hline
\hline
    \multirow{2}{*}{ Model (\text{w/o} BN) } & \multicolumn{3}{c|}{  W8A8} & \multicolumn{3}{c|}{ W4A8} & \multicolumn{3}{c|}{ W2A8} & \multirow{2}[0]{*}{ FP32} \\

\cline{2-10}

     & \multicolumn{1}{c|}{{ ${\mathcal{N}(0,1)}$ }} & \multicolumn{1}{c|}{{ \textsc{ZeroQ}\xspace}\cite{zeroq} } & \multicolumn{1}{c|}{{ \text{GZSQ}}} & \multicolumn{1}{c|}{{ ${\mathcal{N}(0,1)}$ }} & \multicolumn{1}{c|}{{ \textsc{ZeroQ}\xspace} \cite{zeroq}} & \multicolumn{1}{c|}{{ \text{GZSQ}}} & \multicolumn{1}{c|}{{ ${\mathcal{N}(0,1)}$ }} & \multicolumn{1}{c|}{{ \textsc{ZeroQ}\xspace} \cite{zeroq}} & \multicolumn{1}{c|}{{ \text{GZSQ}}} &  \\
\hline    
%\hline

    {Fixup-ResNet 20} & 
    {91.47$\pm$4.6e-3} & 
    \textcolor{blue!95}{{\textbf{91.49$\pm$4.5e-3}}} &
    {91.46$\pm$4.1e-2}
    
    & 91.01$\pm$9.8e-3 
    & {90.97$\pm$6.4e-3} 
    & \textcolor{blue!95}{{\textbf{91.03$\pm$2.7e-2}}} 
    
    & {65.97$\pm$1.8e-2} 
    & 66.00$\pm$1.9e-2 & 
    \textcolor{blue!95}{{\textbf{66.26$\pm$6.4e-2}}} 
    
    & {91.57} \\
    \hline
    
    {Fixup-ResNet 32} & 
    {92.22$\pm$2.1e-2} & 
    {92.21$\pm$0.8e-2} & 
    \textcolor{blue!95}{{\textbf{92.28$\pm$5.1e-2}}} & 
    
    91.83$\pm$6.4e-3 & 
    {91.77$\pm$1.4e-2} & 
    \textcolor{blue!95}{{\textbf{91.95$\pm$3.4e-2}}} & 
    
    {67.19$\pm$1e-13} & 
    \textcolor{blue!95}{{\textbf{67.56$\pm$5.0e-2}}} & 
    {67.46$\pm$9.4e-2} & 
    
    {93.34} \\
    \hline
    
    {Fixup-ResNet 44} & 
    \textcolor{blue!95}{{\textbf{92.20$\pm$8.7e-3}}} & 
    {92.17$\pm$3.4e-2} & 
    {92.20$\pm$5.6e-2} & 
    
    {91.92$\pm$0.0000} & 
    \textcolor{blue!95}{{\textbf{92.00$\pm$1.1e-2}}} & 
    {91.92$\pm$4.6e-2} & 
    
    {71.50$\pm$1.8e-2} & 
    {71.64$\pm$1.7e-2} & 
    \textcolor{blue!95}{{\textbf{71.79$\pm$4.9e-2}}} & 
    
    {92.30} \\
    
    \hline
    {Fixup-ResNet 56} & 
    {92.35$\pm$4.5e-3} & 
    {92.37$\pm$1.0e-2} & 
    \textcolor{blue!95}{{\textbf{92.47$\pm$4.7e-2}}} & 
    
    91.94$\pm$2.4e-2 & 
    {91.89$\pm$1.6e-2} & 
    \textcolor{blue!95}{{\textbf{91.95$\pm$3.7e-2}}} & 
    
    {70.78$\pm$1.6e-2} & 
    70.87$\pm$0.8e-2 & 
    \textcolor{blue!95}{{\textbf{70.89$\pm$6.5e-2}}} & 
    
    {92.61} \\
    
    \hline
    
    {Fixup-ResNet 110} & 
    {92.95$\pm$2.8e-2} & 
    \textcolor{blue!95}{{\textbf{92.97$\pm$2.1e-2}}} & 
    {92.96$\pm$6.4e-2} & 
    
    92.63$\pm$2.3e-2 & 
    {92.53$\pm$6.6e-3} & 
    \textcolor{blue!95}{{\textbf{92.72$\pm$6.4e-2}}} & 
    
    {77.37$\pm$7.4e-3} & 
    77.25$\pm$68e-3 & 
    \textcolor{blue!95}{{\textbf{77.58$\pm$4.0e-2}}} & 
    
    {93.09} \\
%    \hline
\hline
    \end{tabular}}%
%\vspace*{-0.1cm}

  \label{tab:fixup}%
\end{table*}%

\begin{table}[t]
\captionsetup{textfont={sc,footnotesize}, justification=centering, labelsep=newline}
  \centering
\caption{Top-1 accuracy on ISONet \cite{isonet} models for the task of ImageNet \cite{imagenet} classification. ``D'' and ``No-D'' denotes \text{with} and \text{w/o} dropout, respectively. Quantization configuration is INT8 with \text{QS-PC} for weights and QS-PT for activations.}

  \resizebox{0.48\textwidth}{!}{
    \begin{tabular}{|l|c|c|c|c|c|}
    \hline
%\hline
    \multicolumn{2}{|c|}{{ Model (\text{w/o} BN) }} & \multicolumn{1}{c|}{{${\mathcal{N}(0,1)}$}} & \multicolumn{1}{c|}{{ \textsc{ZeroQ}\xspace}\cite{zeroq} } & \multicolumn{1}{c|}{{ \text{GZSQ}}} & \multicolumn{1}{c|}{{ FP32}} \\
    \hline
%\hline

    \multirow{2}[0]{*}{ISONet-18} & {\textcolor{red}{\textbf{No-D}}} & {65.85$\pm$0.11} & {65.86$\pm$0.11} & \textcolor{black}{{\textbf{\textcolor{blue!95}{67.76$\pm$0.02}}}} & {67.94} \\
          & \textcolor{green}{\textbf{D}} & {66.40$\pm$0.09} & {66.41$\pm$0.12} & \textcolor{black}{{\textbf{\textcolor{blue!95}{67.93$\pm$0.01}}}} & {68.10} \\
%          \hline
%   \hline
   \hline       
          
    \multirow{2}[0]{*}{ISONet-34} & {\textcolor{red}{\textbf{No-D}}} & {67.87$\pm$0.12} & {67.85$\pm$0.15} & \textcolor{blue!95}{{\textbf{70.17$\pm$0.03}}} & {70.45} \\
          & \textcolor{green}{\textbf{D}} & {68.97$\pm$0.15} & {68.94$\pm$0.10} & \textcolor{blue!95}{{\textbf{70.75$\pm$0.02}}} & {70.90} \\    
%          \hline
%         \hline
   \hline
     
    \multirow{2}[0]{*}{ISONet-50} & {\textcolor{red}{\textbf{No-D}}} & {66.98$\pm$0.26} & {66.69$\pm$0.26} & \textcolor{blue!95}{{\textbf{70.48$\pm$0.02}}} & {70.73} \\
          & \textcolor{green}{\textbf{D}} & {66.87$\pm$0.28} & {66.90$\pm$0.35} & \textcolor{blue!95}{{\textbf{71.06$\pm$0.03}}} & {71.20} \\
%\hline     

% \hline
   \hline     
          
    \multirow{2}[0]{*}{ISONet-101} & {\textcolor{red}{\textbf{No-D}}} & {66.98$\pm$0.56} & {66.90$\pm$0.60} & \textcolor{blue!95}{{\textbf{70.05$\pm$0.02}}} & {70.38} \\
          & \textcolor{green}{\textbf{D}} & {66.85$\pm$0.25} & {67.29$\pm$0.26} & \textcolor{blue!95}{{\textbf{70.66$\pm$0.02}}} & {71.01} \\
%\hline          
         
%          \hline
   \hline 
          
    \multirow{2}[0]{*}{R-ISONet-18} & \multicolumn{1}{c|}{\textcolor{red}{\textbf{No-D}}} & {64.27$\pm$0.63} & {63.84$\pm$0.14} & \textcolor{blue!95}{{\textbf{68.97$\pm$0.03}}} & {69.06} \\
          & \textcolor{green}{\textbf{D}} & {66.20$\pm$0.23} & {66.10$\pm$0.19} & \textcolor{blue!95}{{\textbf{69.13$\pm$0.03}}} & {69.17} \\
%          \hline
         
%          \hline
   \hline 
          
    \multirow{2}[0]{*}{R-ISONet-34} & {\textcolor{red}{\textbf{No-D}}} & {69.06$\pm$0.21} & {69.08$\pm$0.17} & \textcolor{blue!95}{{\textbf{72.11$\pm$0.03}}} & {72.17} \\
          & \textcolor{green}{\textbf{D}} & {70.94$\pm$0.11} & {70.98$\pm$0.11} & \textcolor{blue!95}{{\textbf{73.33$\pm$0.01}}} & {73.43} \\
%          \hline 
         
%          \hline
   \hline 
          
    \multirow{2}[0]{*}{R-ISONet-50} & \multicolumn{1}{c|}{\textcolor{red}{\textbf{No-D}}} & {70.36$\pm$0.13} & {70.12$\pm$0.18} & \textcolor{blue!95}{{\textbf{74.15$\pm$0.02}}} & {74.20} \\
          & \textcolor{green}{\textbf{D}} & {73.53$\pm$0.15} & {73.56$\pm$0.07} & \textcolor{blue!95}{{\textbf{76.12$\pm$0.01}}} & {76.18} \\
%\hline      

% \hline
   \hline         
          
    \multirow{2}[0]{*}{R-ISONet-101} & \multicolumn{1}{c|}{\textcolor{red}{\textbf{No-D}}} & 73.22$\pm$0.29 & 73.20$\pm$0.29 & \textcolor{blue!95}{{\textbf{75.39$\pm$0.02}}} & 75.44 \\
          & \textcolor{green}{\textbf{D}} & {74.70$\pm$0.12} & {74.79$\pm$0.22} & \textcolor{blue!95}{{\textbf{77.04$\pm$0.03}}} & {77.08} \\
%          \hline
\hline
    \end{tabular}}%
%\vspace*{-0.1cm}

  \label{tab:isonets}%
\end{table}%

\begin{table}[t]
\captionsetup{textfont={sc,footnotesize}, justification=centering, labelsep=newline}
  \centering
    \caption{Top-1 accuracy on various pre-trained models for the task of ImageNet \cite{imagenet} classification. Quantization configuration is INT8 with \text{QS-PC} for weights and QS-PT for activations. }

  \resizebox{0.48\textwidth}{!}{
    \begin{tabular}{|l|c|c|c|c|c|c|c|c|}
%    \hline
\hline
  \multirow{1}[0]{*}{ Model (\text{w} BN)} & \multirow{1}[0]{*}{{ ${\mathcal{N}(0,1)}$ }} &
  \multicolumn{1}{c|}{{ \textcolor{black}{ACIQ} } \cite{NIPS19}}&
   \multicolumn{1}{c|}{{ OCS} \cite{OCS}} &
   \multicolumn{1}{c|}{{\textcolor{black}{DFQ}}\cite{dfq}} & 
   \multicolumn{1}{c|}{{\textcolor{black}{GDFQ}}\cite{gdfq}} & 
   \multicolumn{1}{c|}{{\textsc{ZeroQ}\xspace} \cite{zeroq}} &
%   \multirow{2}[0]{*}{{ GZSQ }} & 
   \multirow{1}[0]{*}{{ \text{GZSQ} }} & \multirow{1}[0]{*}{{  FP32}} \\
%    \hline
%\cmidrule{3-5}
%&&a&\textsc{ICML'19} & 
%%\textbf{ICCV'19} & 
%\textsc{CVPR'20} & 
%%($\mathcal{L}_1 /\mathcal{L}_2$) & 
%&\\
\hline
%Reference & N/A & ICCV'19 & ICML' 19 & CVPR'20 & This work & - \\

Original Data & \textcolor{red}{\ding{55}} &\textcolor{red}{\ding{55}}& \textcolor{green}{\ding{51}} & 
\textcolor{red}{\ding{55}} & \textcolor{red}{\ding{55}} &
\textcolor{red}{\ding{55}} &  \textcolor{red}{\ding{55}} & -\\
%\midrule
%\hline
%\hline
\hline
    {ResNet50} & {77.61$\pm$0.05} & - &76.76& 
%    75.49 & 
    \textcolor{black}{75.62$\pm$0.00}&77.60&{77.60$\pm$0.03} & \textcolor{blue!95}{{\textbf{77.65$\pm$0.05}}} & {77.72} \\
    \hline
    {MobileNetV2} & \textcolor{blue!95}{{\textbf{72.93$\pm$0.02}}} & \textcolor{black}{71.75}  &-&
%    71.20 &
    \textcolor{black}{71.20$\pm$0.00}&72.80&72.92$\pm$0.00 & \textcolor{blue!95}{{\textbf{72.93$\pm$0.02}}} & {73.03} \\
    \hline
    {ShuffleNetV1} & {64.45$\pm$0.04} & \textcolor{black}{63.49} &-&
%    - &
    -&-&{64.17$\pm$0.13} & \textcolor{blue!95}{{\textbf{64.66$\pm$0.06}}} & {65.07} \\
    \hline
    {ResNet18} & {69.90$\pm$0.16} & \textcolor{black}{70.77} &71.35&
%    69.26 &
    \textcolor{black}{71.00$\pm$0.00}&70.74&71.38$\pm$0.02 & \textcolor{blue!95}{{\textbf{71.44$\pm$0.01}}} & {71.47} \\
    \hline
    {ResNet26} & {72.25$\pm$0.10} & - &73.14& 
%    72.75 & 
    \textcolor{black}{72.48$\pm$0.00}&\textcolor{black}{72.69}&{73.57$\pm$0.04} & \textcolor{blue!95}{{\textbf{73.59$\pm$0.02}}} & {73.70} \\
    \hline
    {ResNet152} & {78.79$\pm$0.05} & - &78.58& 
%    77.78 &
    \textcolor{black}{78.66$\pm$0.00}&78.54&\textcolor{blue!95}{{\textbf{78.90$\pm$0.04}}} & {78.85$\pm$0.02} & {80.08} \\
    \hline
    {InceptionV3} & \textcolor{blue!95}{{\textbf{77.82$\pm$0.13}}} & \textcolor{black}{78.73} &76.68& 
%    - &
    -&-&78.79$\pm$0.04 & 78.79$\pm$0.03 & {78.88} \\
    \hline
    {SqueezeNextV5} & {66.99$\pm$0.17} & \textcolor{black}{68.99} &63.20&
%    - &
    -&-&{68.06$\pm$0.21} & \textcolor{blue!95}{{\textbf{69.32$\pm$0.05}}} & {69.38} \\
\hline    
    {MobileNetV1} & {72.15$\pm$0.10}& - &-& 
%    -&
    -&-&{72.11$\pm$0.21} & \textcolor{blue!95}{{\textbf{72.99$\pm$0.04}}} & {73.39} \\
%    \hline
%\midrule
%    \textcolor{black}{RetinaNet} & \textcolor{black}{0.352} & {\textbf{0.363}} & {\textbf{0.363}} & \textcolor{black}{0.364} \\
\hline
%\bottomrule
    \end{tabular}}%
  \label{tab:zeroq-imagenet}%
\end{table}%

\begin{table}[h]
\captionsetup{textfont={sc,footnotesize}, justification=centering, labelsep=newline}
\centering
\caption{Mean Top-1 accuracy on MobileNetV2 \cite{mobilenetv2} for ImageNet \cite{imagenet} classification. Quantization configuration is INT8. No D: Data-free. No FT: No Fine-tuning. No AN: No Additional Network. No BN \cite{bn}: Works w/o BN \cite{bn} layers.
}
\begin{tabular}{|l|c|c|c|c|c|}
\hline 
Method & No D & No FT & No AN & No BN \cite{bn} & Top-1\\
\hline
%\hline
FP32 & - & - & - & - &  73.03\\
\hline
RVQuant \cite{rvquant} &  \textcolor{red}{\ding{55}} & \textcolor{red}{\ding{55}} & \textcolor{green}{\ding{51}} &
 \textcolor{green}{\ding{51}} &  70.29\\
 \hline
\textcolor{black}{ACIQ} \cite{NIPS19} &  \textcolor{red}{\ding{55}} & \textcolor{green}{\ding{51}} &\textcolor{green}{\ding{51}} & \textcolor{green}{\ding{51}} &  70.68\\
\hline
DFQ \cite{dfq} &  \textcolor{green}{\ding{51}} & \textcolor{green}{\ding{51}} & \textcolor{green}{\ding{51}} & \textcolor{red}{\ding{55}} &  71.20\\
\hline
GDFQ \cite{gdfq} &  \textcolor{green}{\ding{51}} & \textcolor{green}{\ding{51}} & \textcolor{red}{\ding{55}} &\textcolor{red}{\ding{55}} & 72.80 \\
\hline
\textsc{ZeroQ} \cite{zeroq} &  \textcolor{green}{\ding{51}} & \textcolor{green}{\ding{51}} &
\textcolor{green}{\ding{51}} &
 \textcolor{red}{\ding{55}} &  72.92\\
 \hline
GZSQ &  \textcolor{green}{\ding{51}} & \textcolor{green}{\ding{51}} & \textcolor{green}{\ding{51}} &\textcolor{green}{\ding{51}} &  \textbf{\textcolor{blue!95}{72.93}}\\
\hline
\end{tabular}
\label{tab:mix_results_1}
\end{table}

\begin{table}[h]
\captionsetup{textfont={sc,footnotesize}, justification=centering, labelsep=newline}
\centering
\caption{Mean Top-1 accuracy on ResNet18 \cite{resnet} for ImageNet \cite{imagenet} classification. Quantization configuration is INT8. No D: Data-free. No FT: No Fine-tuning. No AN: No Additional Network. No BN \cite{bn}: Works w/o BN \cite{bn} layers.}
\begin{tabular}{|l|c|c|c|c|c|}
\hline 
{Method} &  {No D} & {No FT} & {No AN} & {No BN} \cite{bn}  & {Top-1}\\
\hline
FP32 &  - &- & - & - &  71.47\\
\hline
MSE \cite{MSE_Quant} &  \textcolor{red}{\ding{55}} & \textcolor{red}{\ding{55}} & \textcolor{green}{\ding{51}} & \textcolor{green}{\ding{51}} &  71.01\\
\hline
KL \cite{nips16_w} &  - & - &- & - &  70.69\\
\hline
RVQuant \cite{rvquant} &  \textcolor{red}{\ding{55}} & \textcolor{red}{\ding{55}} &\textcolor{green}{\ding{51}} & \textcolor{green}{\ding{51}} &  70.01\\
\hline
OCS \cite{OCS} &  \textcolor{red}{\ding{55}} & \textcolor{green}{\ding{51}} &\textcolor{green}{\ding{51}} & \textcolor{green}{\ding{51}} &  71.35\\
\hline
ACIQ \cite{NIPS19} &  \textcolor{red}{\ding{55}} & \textcolor{green}{\ding{51}} &\textcolor{green}{\ding{51}} & \textcolor{green}{\ding{51}} &  68.78\\
\hline
DFQ \cite{dfq} &  \textcolor{green}{\ding{51}} & \textcolor{green}{\ding{51}} &\textcolor{green}{\ding{51}} & \textcolor{red}{\ding{55}} &  69.70\\
\hline
DFC \cite{knowledge-within} &  \textcolor{green}{\ding{51}} & \textcolor{red}{\ding{55}} & \textcolor{green}{\ding{51}} & \textcolor{red}{\ding{55}} &  69.57\\
\hline
GDFQ \cite{gdfq} &  \textcolor{green}{\ding{51}} & \textcolor{green}{\ding{51}} & \textcolor{red}{\ding{55}} & \textcolor{red}{\ding{55}} &  71.43\\
\hline
\textsc{ZeroQ} \cite{zeroq} &  \textcolor{green}{\ding{51}} & \textcolor{green}{\ding{51}} &
\textcolor{green}{\ding{51}} & \textcolor{red}{\ding{55}} &  71.38\\
\hline
GZSQ &  \textcolor{green}{\ding{51}} & \textcolor{green}{\ding{51}} & \textcolor{green}{\ding{51}} & \textcolor{green}{\ding{51}} &  \textbf{\textcolor{blue!95}{71.44}}\\
\hline
\end{tabular}
\label{tab:mix_results_2}
\end{table}

\begin{table*}[t]
\captionsetup{textfont={sc,footnotesize}, justification=centering, labelsep=newline}
  \centering
      \caption{Top-1 accuracy on various models for the ImageNet \cite{imagenet} classification. Quantization configuration is INT8 with \text{QS-PT} for both weights and activations. $^*$ Used 1K samples (1/class) from ImageNet \cite{imagenet} training set for range calibration of the activations.}
%\setlength{\aboverulesep}{0pt}
%\setlength{\belowrulesep}{0pt}
%  \caption{Add caption- Per channel symmetric}
\resizebox{\textwidth}{!}{
    \begin{tabular}{|l|c|c|c|c|c|c|c|c|c|}
%    \hline
\hline

    \multirow{2}[0]{*}{ Model (\text{w} BN) } & \multicolumn{2}{c|}{ ${\mathcal{N}(0,1)}$} & \multicolumn{2}{c|}{{ \textsc{ZeroQ}\xspace}\cite{zeroq} } & \multicolumn{2}{c|}{ \text{GZSQ}} & \multicolumn{2}{c|}{ $\text{Train}^*$} & \multicolumn{1}{c|}{\multirow{2}[0]{*}{ FP32}} \\
    \cline{2-9}
    \multicolumn{1}{|c|}{} & \multicolumn{1}{c|}{ Before} & \multicolumn{1}{c|}{ After} & \multicolumn{1}{c|}{ Before} & \multicolumn{1}{c|}{After} & \multicolumn{1}{c|}{ Before} & \multicolumn{1}{c|}{ After} & \multicolumn{1}{c|}{Before} & \multicolumn{1}{c|}{After} &  \\
\hline
%\hline    
    
    ResNet18 & 67.42$\pm$0.23 & 67.78$\pm$0.26 & 67.50$\pm$0.21 & 15.97$\pm$1.16 & \textcolor{blue!95}{\textbf{69.27$\pm$0.05}} &\textcolor{blue!95}{\textbf{69.09$\pm$0.05}} & 69.37 & 69.29 & 69.76 \\
    \hline
    MobileNetV2 & 68.47$\pm$0.10 & 68.45$\pm$0.17& \textcolor{blue!95}{\textbf{68.71$\pm$0.14}} & 51.58$\pm$4.13  & 68.50$\pm$0.18 & \textcolor{blue!95}{\textbf{68.55$\pm$0.17}} & 67.21 & 67.08 & 71.88 \\
    \hline
    ShuffleNetV2 & 62.47$\pm$0.13 & 62.38$\pm$0.11 & 62.34$\pm$0.20 & 58.27$\pm$1.11 & \textcolor{blue!95}{\textbf{64.04$\pm$0.12}} & \textcolor{blue!95}{\textbf{64.02$\pm$0.18}} & 64.74 & 64.13 & 69.36 \\
    \hline
    ResNet50 & 75.27$\pm$0.04 & 75.22$\pm$0.08 & 75.27$\pm$0.04 & 14.69$\pm$1.91 & \textcolor{blue!95}{\textbf{75.40$\pm$0.04}} & \textcolor{blue!95}{\textbf{75.27$\pm$0.12}} & 75.34 & 75.32 & 76.15 \\
\hline    
    InceptionV3 & 76.15$\pm$0.04 & \textcolor{blue!95}{\textbf{76.16$\pm$0.03}} & \textcolor{blue!95}{\textbf{76.17$\pm$0.04}} & 73.04$\pm$2.82 & 76.15$\pm$0.03  & 76.16$\pm$0.04 & 76.27 & 76.19  & 77.45 \\
%    \hline
\hline   
    \end{tabular}}%
%\vspace*{-0.1cm}
  \label{tab:pytorch_quant_qspt}%
\end{table*}%

\begin{table*}[t]
\captionsetup{textfont={sc,footnotesize}, justification=centering, labelsep=newline}
  \centering
      \caption{Top-1 accuracy on various models for the ImageNet \cite{imagenet} classification. Quantization configuration is INT8 with \text{QS-PC} for weights, and QS-PT for activations. $^*$ Used 1K samples (1/class) from ImageNet \cite{imagenet} training set for range calibration of the activations.}
%\setlength{\aboverulesep}{0pt}
%\setlength{\belowrulesep}{0pt}
%  \caption{Add caption- Per channel symmetric}
\resizebox{\textwidth}{!}{
    \begin{tabular}{|l|c|c|c|c|c|c|c|c|c|}
%    \hline
\hline
    \multirow{2}[0]{*}{ Model (\text{w} BN) } & \multicolumn{2}{c|}{ ${\mathcal{N}(0,1)}$} & \multicolumn{2}{c|}{{ \textsc{ZeroQ}\xspace}\cite{zeroq} } & \multicolumn{2}{c|}{ \text{GZSQ}} & \multicolumn{2}{c|}{ $\text{Train}^*$} & \multicolumn{1}{c|}{\multirow{2}[0]{*}{ FP32}} \\
    \cline{2-9}
    \multicolumn{1}{|c|}{} & \multicolumn{1}{c|}{ Before} & \multicolumn{1}{c|}{ After} & \multicolumn{1}{c|}{ Before} & \multicolumn{1}{c|}{After} & \multicolumn{1}{c|}{ Before} & \multicolumn{1}{c|}{ After} & \multicolumn{1}{c|}{Before} & \multicolumn{1}{c|}{After} &  \\
\hline
%\hline

    ResNet18 &67.80$\pm$0.37&67.78$\pm$0.19&67.76$\pm$0.17&16.17$\pm$2.13&\textbf{\textcolor{blue!95}{69.47$\pm$0.05}}&\textbf{\textcolor{blue!95}{69.33$\pm$0.06}}&69.43&69.44&69.76 \\
    \hline
    MobileNetV2 &69.40$\pm$0.13&69.32$\pm$0.10&\textbf{\textcolor{blue!95}{69.80$\pm$0.12}}&54.94$\pm$1.61&69.38$\pm$0.10&\textbf{\textcolor{blue!95}{69.58$\pm$0.11}}&67.99&67.15&71.88 \\
    \hline
    ShuffleNetV2  &67.23$\pm$0.13&67.25$\pm$0.17&67.51$\pm$0.13&58.60$\pm$2.21&\textbf{\textcolor{blue!95}{67.72$\pm$0.08}}&\textbf{\textcolor{blue!95}{67.88$\pm$0.14}}&68.15&68.17&69.36 \\
    \hline
    ResNet50  &75.61$\pm$0.03&75.60$\pm$0.05&75.72$\pm$0.05&13.24$\pm$0.87&\textbf{\textcolor{blue!95}{75.83$\pm$0.04}}&\textbf{\textcolor{blue!95}{75.61$\pm$0.12}}&75.94&75.79&76.15 \\
\hline    
    InceptionV3  &77.03$\pm$0.02&77.02$\pm$0.03&\textbf{\textcolor{blue!95}{77.16$\pm$0.05}}&72.49$\pm$1.89&77.09$\pm$0.01&\textbf{\textcolor{blue!95}{77.14$\pm$0.05}}&77.16&77.20&77.45 \\
%    \hline
\hline   
    \end{tabular}}%
%\vspace*{-0.1cm}

  \label{tab:pytorch_quant_qspc}%
\end{table*}%

\section{Results}
\label{results}
\subsection{Classification}

Table~\ref{tab:fixup} shows the CIFAR-10 classification accuracy on Fixup-ResNets \cite{fixup} models that are w/o BN \cite{bn} layers, when quantized with different configuration. W8A8 denotes 8-bit integer precision for both weights and activations. It can be observed that for most of the Fixup-ResNet variants, the proposed GZSQ has outperformed the existing \textsc{ZeroQ} \cite{zeroq} approach across different quantization configurations, especially for W2A8. It should also be noted that for W2A8\footnote{considering the significant difference in the bit-precisions of 2-bits for weights and 8-bits for activations}, even though the proposed GZSQ utilizes the pre-trained weights for estimating the $\Delta$ and $z$ of the activations, there is not much degradation in the accuracy, compared to the \textsc{ZeroQ} \cite{zeroq}.

Similarly, Table~\ref{tab:isonets} shows the quantization accuracy of ISONets \cite{isonet} models that are also w/o BN \cite{bn} layers on ImageNet \cite{imagenet} classification. It can be observed that for most of the models, GZSQ has achieved near FP32 accuracy, outperforming the \textsc{ZeroQ} \cite{zeroq} by a significant margin. From Tables~\ref{tab:fixup}, \ref{tab:isonets}, it may be noted that \textsc{ZeroQ}\xspace \cite{zeroq} behaves similar to unit Gaussian in the absence of BN \cite{bn} layers due to its underlying backbone architecture, and results lower accuracy with quantized models. Whereas, the results obtained by using proposed \textsc{GZSQ}\xspace are not only superior to unit Gaussian and \textsc{ZeroQ}\xspace \cite{zeroq}, but also close to FP32 accuracy. 
%It should also be noted from Tables~\ref{tab:fixup}, \ref{tab:isonets} that in the absence of BN layers, \textsc{ZeroQ} \cite{zeroq} behaves similar to unit Gaussian ($\mathcal{N}(0,1)$).
%We have also shown the results for the models that do not use BN layers, such as ISONets \cite{isonet}, and Fixup-ResNets \cite{fixup}, in Tables~\ref{tab:isonets}, \ref{tab:fixup}, respectively. 

Table~\ref{tab:zeroq-imagenet} shows the results obtained on ImageNet \cite{imagenet} classification for various models\footnote{Available: \url{https://pypi.org/project/pytorchcv/}} that consists of BN \cite{bn} layers. It can be observed that the proposed \textsc{GZSQ}\xspace has shown a remarkable improvement of over OCS \cite{OCS} and \textsc{ZeroQ}\xspace \cite{zeroq} on majority of the models, especially with $+$1.25\% for SqueezeNextV5 \cite{nndesign1}. Tables. \ref{tab:mix_results_1}, \ref{tab:mix_results_2} present the ImageNet \cite{imagenet} classification accuracy for MobileNetV2 \cite{mobilenetv2} and ResNet18 \cite{resnet} models, respectively. The results are categorized based on the methods that do not (a) No D: require original unlabelled samples, (b) No FT: require finetuning, (c) No AN: require additional network, and (d) No BN: require BN\cite{bn} layers in the network.
%We have also given a quantitative study to show the comparative improvement of $\mathcal{L}_\mathbf{D}$ over $\mathcal{L}_1$ and $\mathcal{L}_2$ norms in Section~\ref{cost_fn_ablation}. Note that the models evaluated in Table~\ref{tab:zeroq-imagenet} comprise of BN \cite{bn} layers. 

The results in Tables~\ref{tab:fixup}:\ref{tab:zeroq-imagenet},  \ref{tab:loss_ablation} are computed using \textsc{ZeroQ}\xspace 's \cite{zeroq} quantization simulation, whereas the results shown in  Tables~\ref{tab:pytorch_quant_qspt}:\ref{tab:size},  \ref{tab:bias_vs_no_bias} are computed using Pytorch's post-training static quantization framework. It can be observed from the Tables~\ref{tab:pytorch_quant_qspt}, \ref{tab:pytorch_quant_qspc} that the proposed \textsc{GZSQ}\xspace has outperformed the \textsc{ZeroQ}\xspace \cite{zeroq} on majority of the models\footnote{{Source: \url{https://github.com/pytorch/vision/tree/master/torchvision/models/quantization}}}. Even though we have assumed the input for the data distillation (see Sections \ref{sec:se}, \ref{sec:esa}, \ref{sec:dd}) to be a random $\mathcal{N}(0,1)$ instance, the \textsc{GZSQ}\xspace has shown a significant performance gain over Gaussian (0,1) input. It may be due to the incorporation of pre-trained weights in the proposed distillation process that has not been considered in \textsc{ZeroQ}\xspace \cite{zeroq}. 

{It may also be concluded that the \textsc{GZSQ}\xspace 's distilled data has been more beneficial for inferring the activation ranges over $\mathcal{N}(0,1)$ and \textsc{ZeroQ}\xspace 's \cite{zeroq}.} We have shown the results for both the cases when BN \cite{bn} folding is performed \text{before} and \text{after} data distillation. We have also shown the quantization results when a random subset of training set is used to infer the activation ranges. 

\subsection{Object Detection}
\textcolor{black}{Among high-level vision tasks, the complexity increases when attention shifts from classification to object detection in an image. Object detection aims to predict the set of bounding boxes around the objects with pre-defined category labels in an input image. To demonstrate the robustness of our approach,} we have also evaluated the proposed scheme on the task of object detection using the Microsoft COCO \cite{coco} dataset. \textcolor{black}{For this,} we have adopted RetinaNet \cite{retina} \textcolor{black}{single-stage detector} which has a state-of-the-art results i.e., 36.4 mean average precision (mAP) (\textcolor{black}{\texttt{standard 0.5:0.05:0.95 metric}}). \textcolor{black}{The adopted RetinaNet has a ResNet50 as backbone architecture. For quantization, in this paper, we have processed the pre-trained RetinaNet with INT8 configuration by using QS-PC for weights and QS-PT for activations. It has been observed that} while both \textsc{ZeroQ}\xspace \cite{zeroq} and \textsc{GZSQ}\xspace result in 36.3 mAP, $\mathcal{N}(0,1)$ result in 35.2 mAP. \textcolor{black}{Although the GSZQ obtained similar results to \textsc{ZeroQ}, it can be concluded that the proposed GZSQ is also flexible to be used across different problem domains. }

\begin{figure}[t]
\centering
\hspace*{-0.45cm}
\setlength{\tabcolsep}{1pt}
\begin{tabular}{cc}
\includegraphics[width=0.22\textwidth]{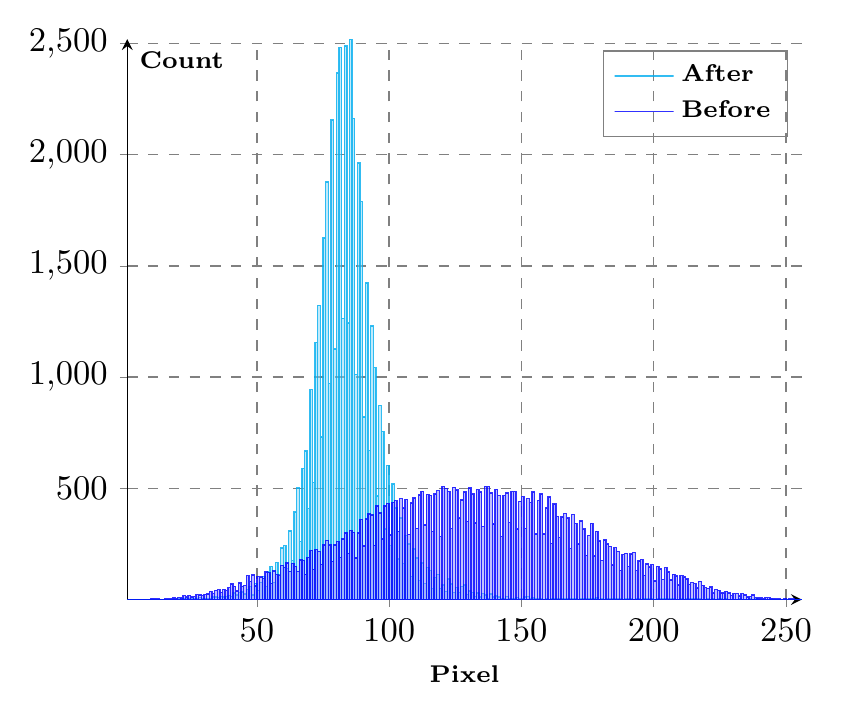}&
\includegraphics[width=0.22\textwidth]{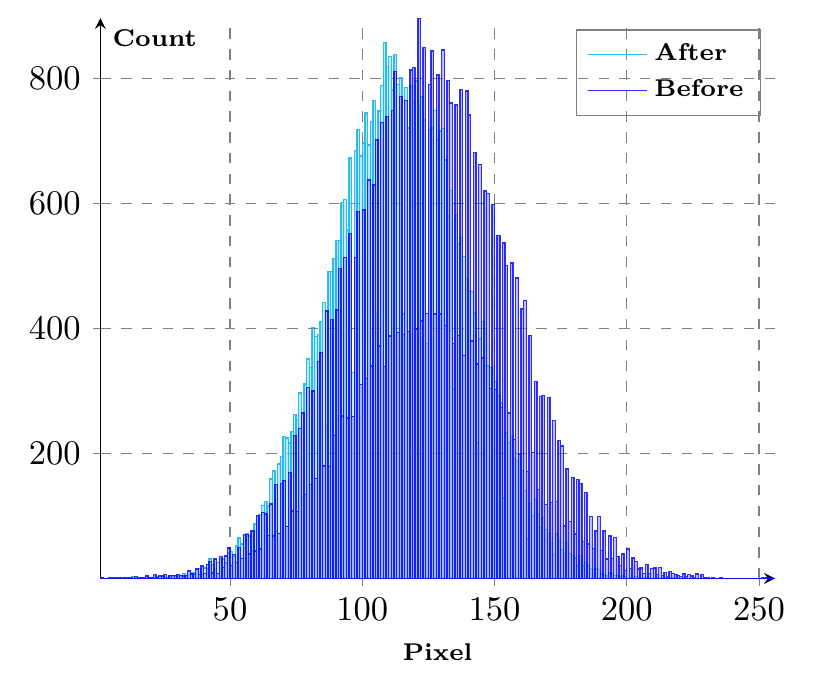}\\
\textsc{ZeroQ} \cite{zeroq} & GZSQ\\
\end{tabular}
\caption{Samples histograms of distilled data for the adopted medical imaging task on ResNext101 \cite{resnext}.}
\label{fig:dd_2}
\end{figure}
\begin{figure*}[t]
\centering
\hspace*{-0.5cm}
\setlength{\tabcolsep}{2pt}
\begin{tabular}{ccc}

\includegraphics[width=0.3\textwidth]{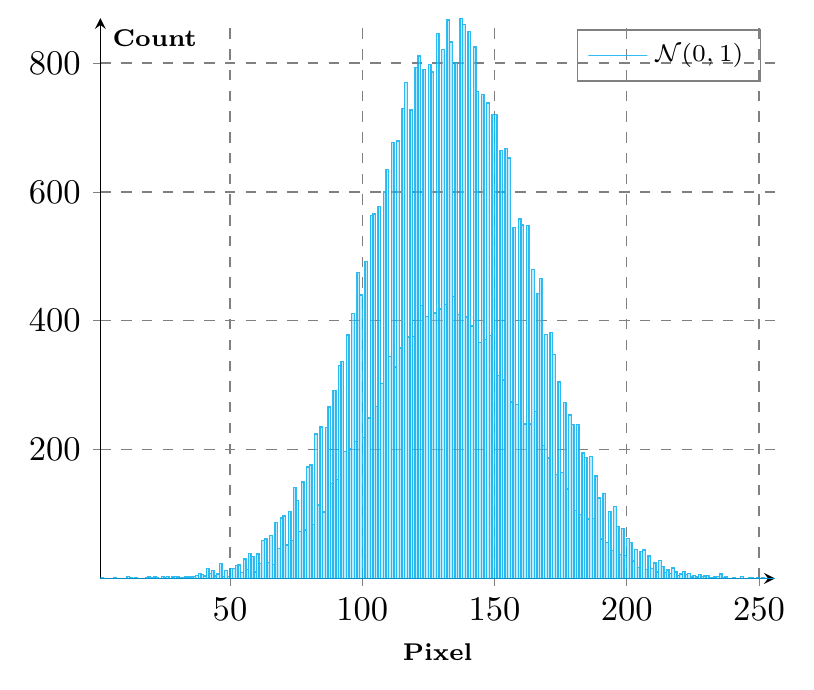}
&
\includegraphics[width=0.3\textwidth]{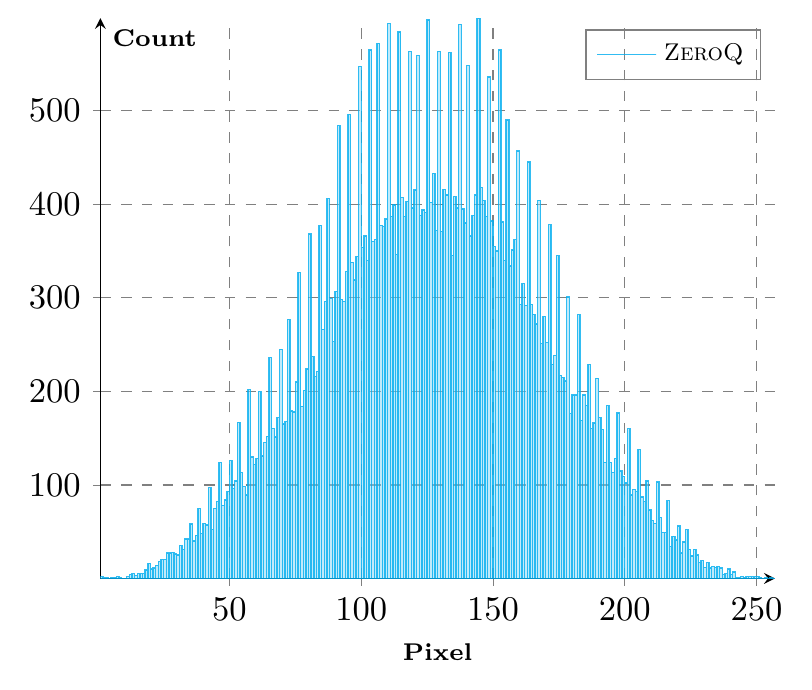}
&
\includegraphics[width=0.3\textwidth]{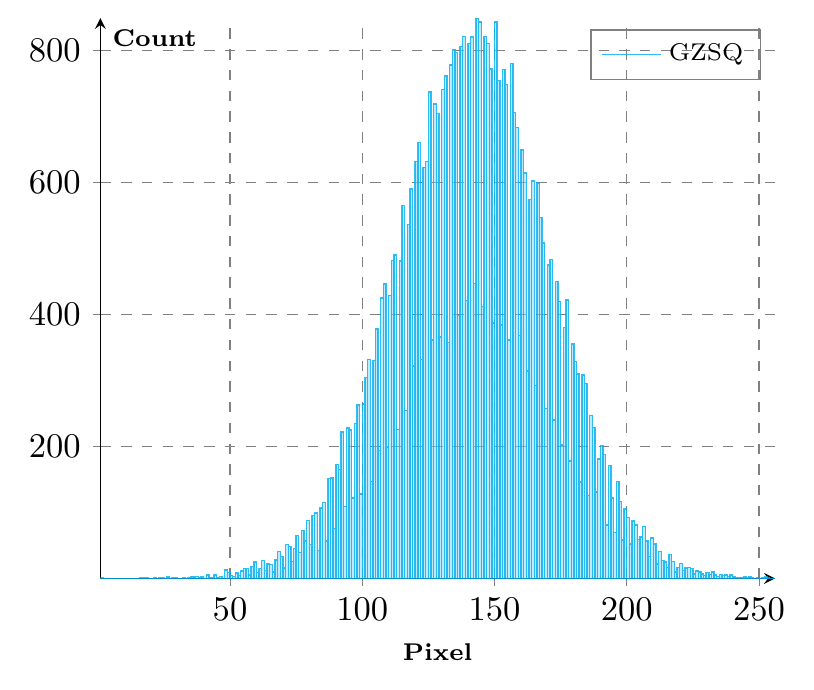}
\\

\includegraphics[width=0.3\textwidth]{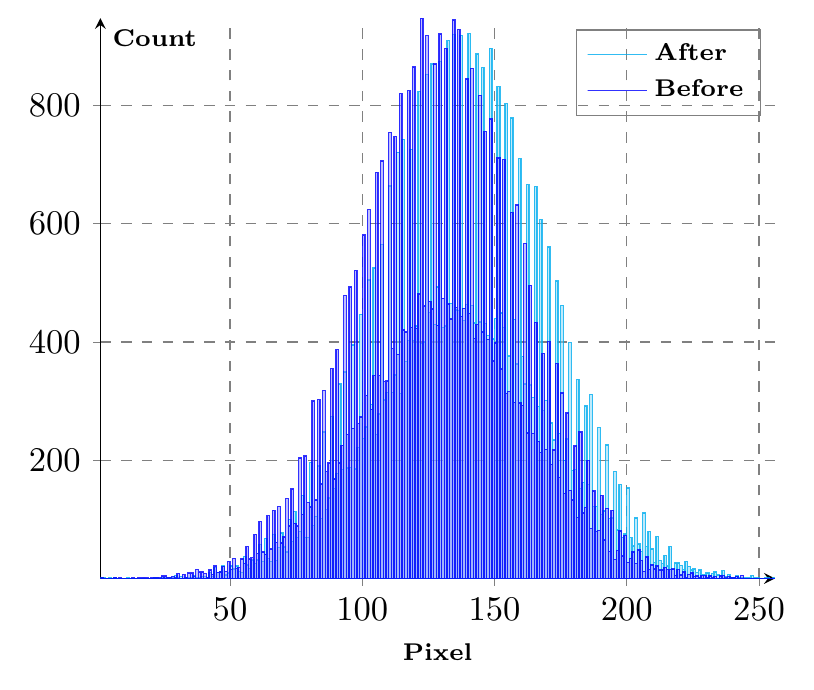}
&
\includegraphics[width=0.3\textwidth]{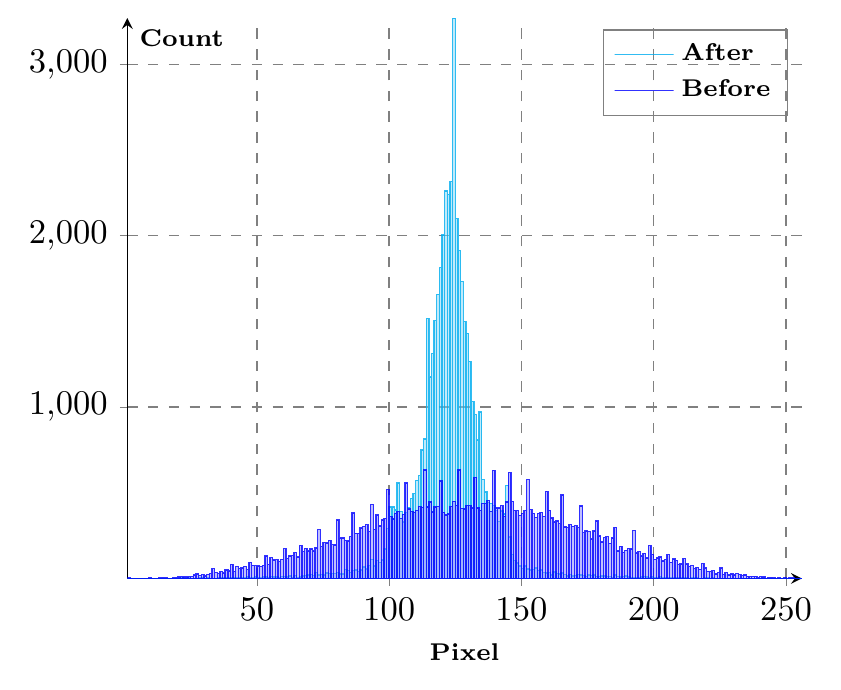}
&
\includegraphics[width=0.3\textwidth]{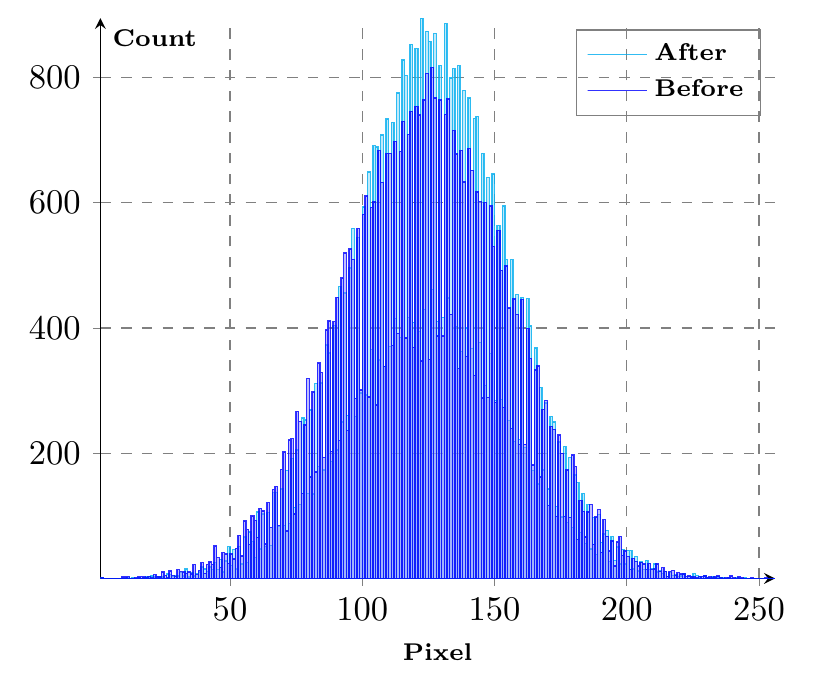}
\\
$\mathcal{N}(0,1)$ & \textsc{ZeroQ} \cite{zeroq} & GZSQ\\
\end{tabular}
\caption{Sample histograms of the distilled data generated by using $\mathcal{N}(0, 1)$, \textsc{ZeroQ} \cite{zeroq}, and the proposed GZSQ schemes. {Row 1} shows the data for the ISONet \cite{isonet} model. {Row 2} shows the data for MobileNetV2 \cite{mobilenetv2} when BN-folding is performed before and after data distillation. }
\label{fig:dd_1}
\end{figure*}

\begin{figure*}[h]
\centering
\setlength{\tabcolsep}{2pt}
\begin{tabular}{cccc}

\begin{overpic}[width=0.23\textwidth]{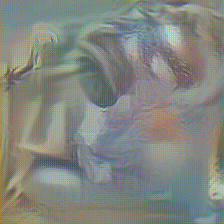}
     \put(35,1){\includegraphics[width=0.15\textwidth]{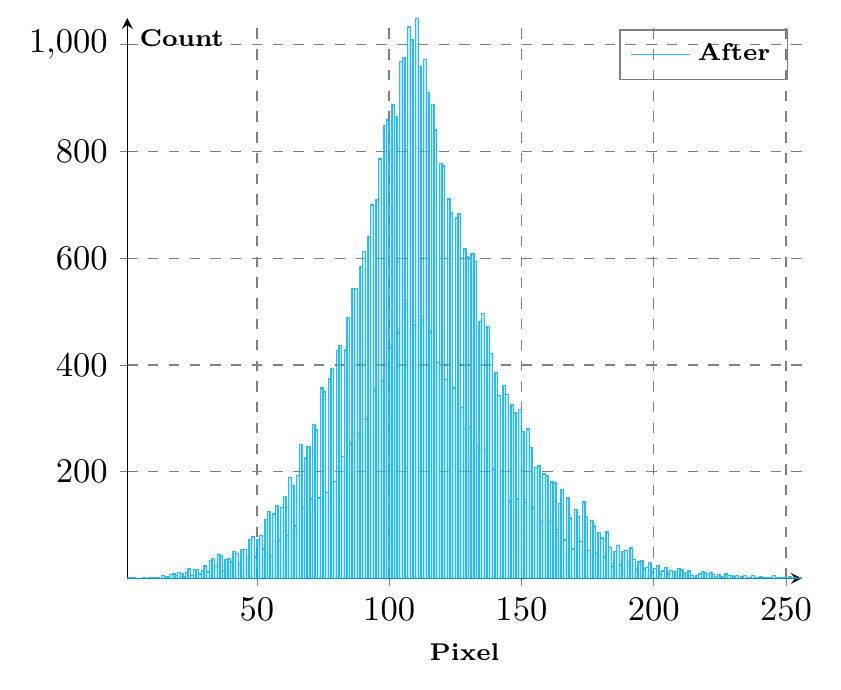}}
  \end{overpic}
&
\begin{overpic}[width=0.23\textwidth]{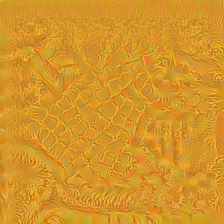}
     \put(35,1){\includegraphics[width=0.15\textwidth]{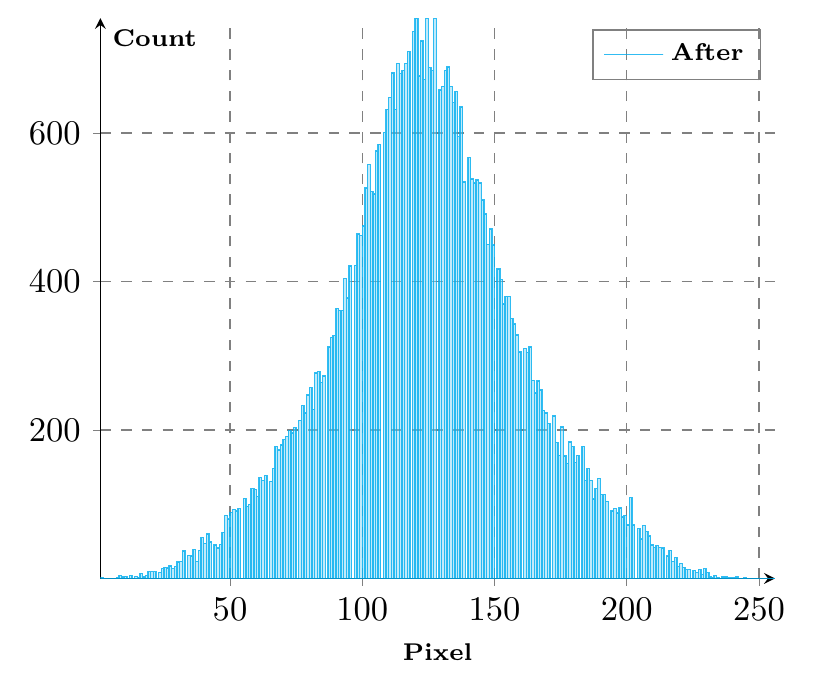}}
  \end{overpic}
  &
  \begin{overpic}[width=0.23\textwidth]{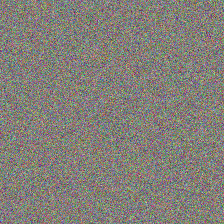}
     \put(35,1){\includegraphics[width=0.15\textwidth]{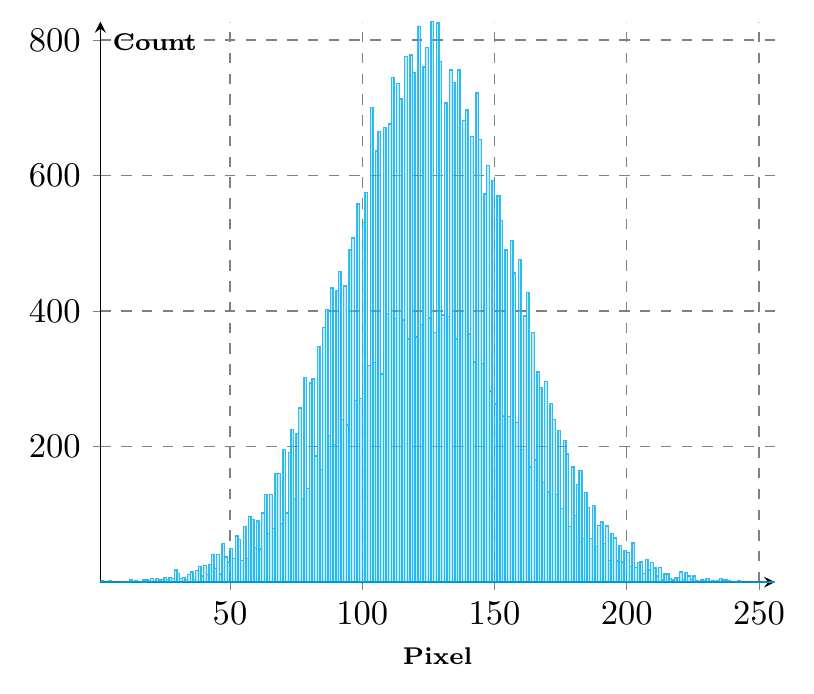}}
  \end{overpic}
  &
  \begin{overpic}[width=0.23\textwidth]{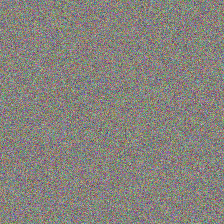}
     \put(35,1){\includegraphics[width=0.15\textwidth]{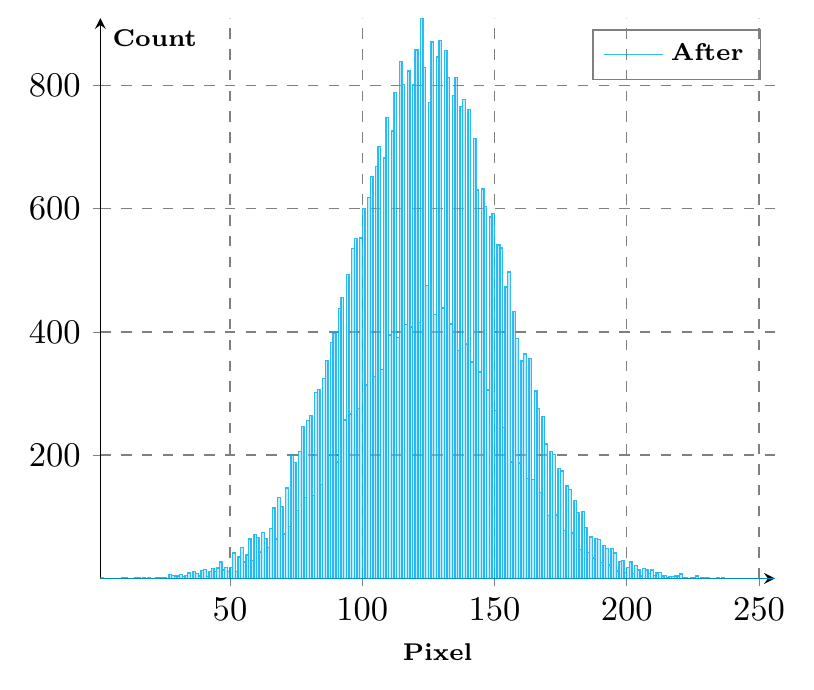}}
  \end{overpic}\\
  
  \begin{overpic}[width=0.23\textwidth]{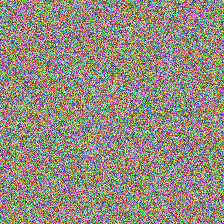}
     \put(35,1){\includegraphics[width=0.15\textwidth]{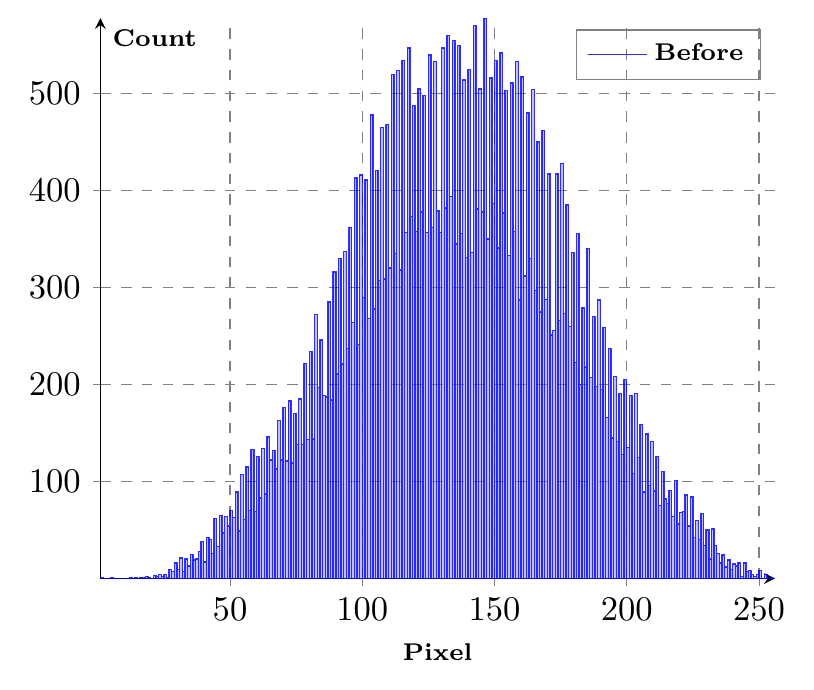}}
  \end{overpic}
&
\begin{overpic}[width=0.23\textwidth]{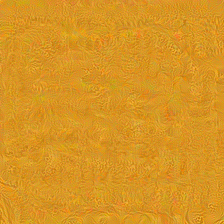}
     \put(35,1){\includegraphics[width=0.15\textwidth]{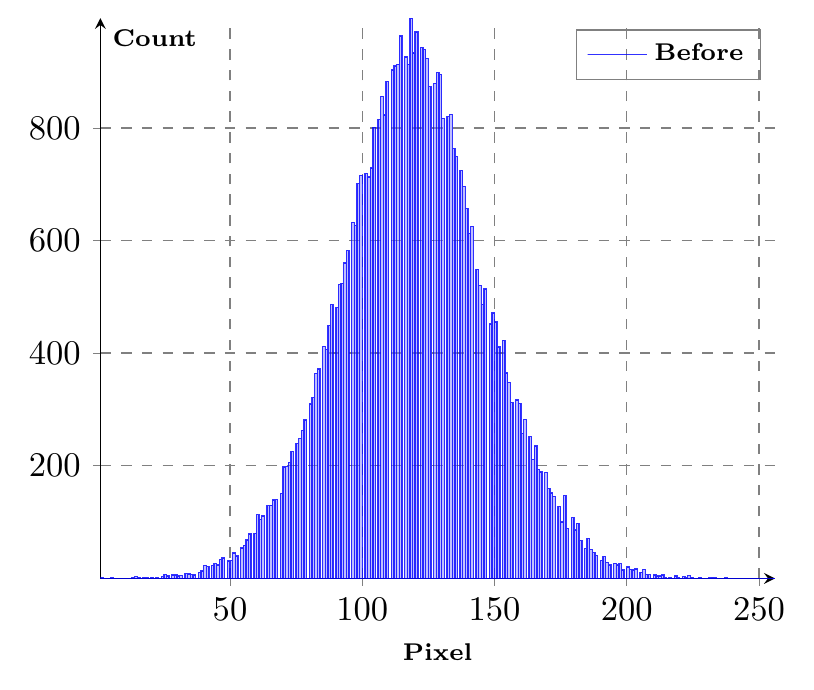}}
  \end{overpic}
  &
  \begin{overpic}[width=0.23\textwidth]{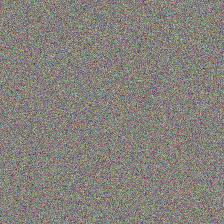}
     \put(35,1){\includegraphics[width=0.15\textwidth]{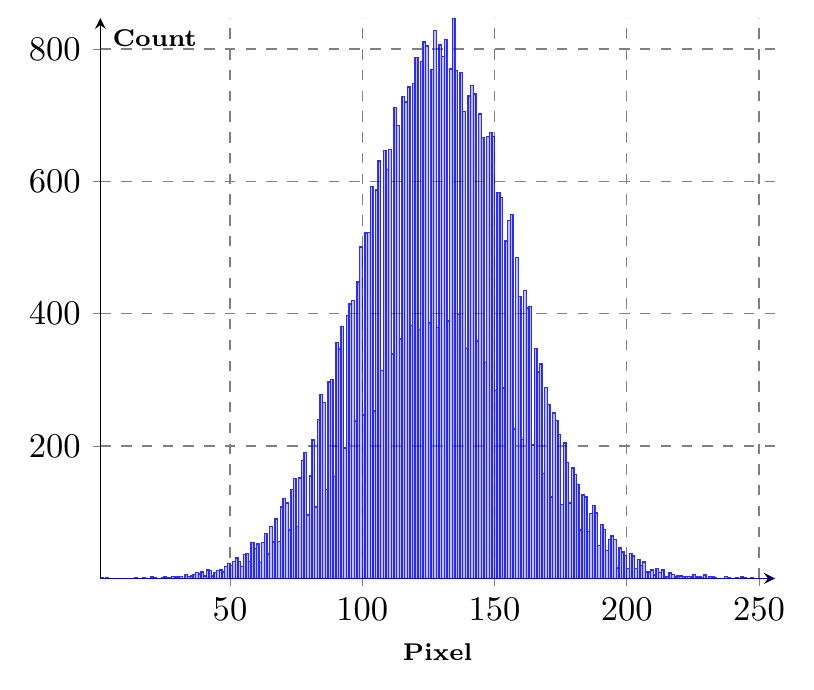}}
  \end{overpic}
  &
  \begin{overpic}[width=0.23\textwidth]{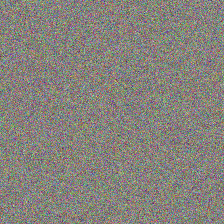}
     \put(35,1){\includegraphics[width=0.15\textwidth]{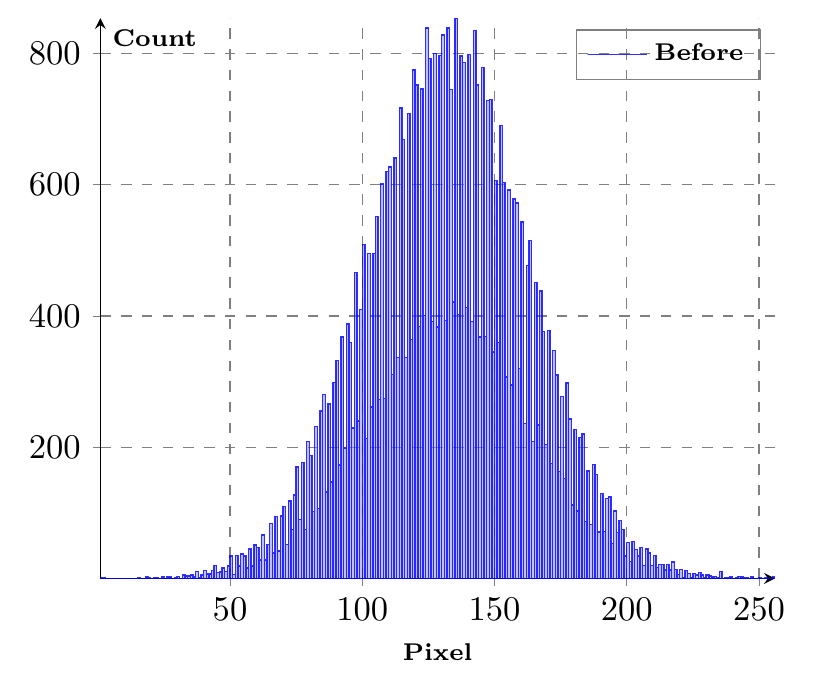}}
  \end{overpic}\\

\textsc{ZeroQ} \cite{zeroq} & GZSQ-$\ell_2$ & GZSQ-$\mathcal{L}^{*}_{KL}$ & GZSQ\\

\end{tabular}
\caption{Sample stills and corresponding histograms of the distilled data when BN \cite{bn} layer is folded before (Row-bottom) and after (Row-top) distillation on ResNet18 \cite{resnet} for the task of ImageNet \cite{imagenet} classification.}
\label{fig:cf_1}
\end{figure*}

\subsection{Medical Imaging}
Unlike natural images, X-rays and Magnetic Resonance Imaging (MRI) \cite{mri} scans do not comprise of many indistinguishable features, in general, among each other. It is challenging to differentiate between the X-rays or MRI scans of a normal and diseased person. Several methods based on deep learning frameworks have been proposed for many diseases \cite{med1, med2}. In this work, considering the COVID-19 \cite{covid19} era, we have adopted one of the significant problems, namely, Pneumonia classification \cite{medical1} in chest X-rays. For this, we have considered the ResNet18 \cite{resnet} and ResNext101 \cite{resnext} as baseline models and trained for the binary classification problem using the benchmark dataset \cite{pneumonia} with a pre-determined train/val/test split. The trained model then has been quantized, and 
\begin{table}[t]
\captionsetup{textfont={sc,footnotesize}, justification=centering, labelsep=newline}
  \centering
  \caption{Top-1 accuracy of the quantized models for Pneumonia classification in chest X-Rays. }
    \begin{tabular}{|l|cc|c|}
    \hline
    \multicolumn{2}{|c|}{Methods} & \multicolumn{1}{c|}{ResNet18} & \multicolumn{1}{c|}{ResNext101\_32x8d} \\
    \hline
%\hline    
    
    \multirow{2}[0]{*}{$\mathcal{N}(0,1)$} & \multicolumn{1}{c|}{Before} & 80.40$\pm$0.23      & 80.81$\pm$0.44 \\
          & \multicolumn{1}{c|}{After} &   80.36$\pm$0.12    & 80.89$\pm$0.47 \\
          \cline{1-4}
    \multirow{2}[0]{*}{\textsc{ZeroQ} \cite{zeroq}} & \multicolumn{1}{c|}{Before} &    80.40$\pm$0.14   & 81.04$\pm$0.35 \\
          & \multicolumn{1}{c|}{After} &   79.43$\pm$0.44    & 79.83$\pm$0.31 \\
          \cline{1-4}
    \multirow{2}[0]{*}{\text{GZSQ}} & \multicolumn{1}{c|}{Before} &   \textbf{\textcolor{blue!95}{80.56$\pm$0.36}}    & \textbf{\textcolor{blue!95}{81.16$\pm$0.53}} \\
          & \multicolumn{1}{c|}{After} &  \textbf{\textcolor{blue!95}{80.48$\pm$0.15}}    & \textbf{\textcolor{blue!95}{81.02$\pm$0.48}}  \\
          \cline{1-4}
    \multicolumn{2}{|c|}{FP32} &    80.80   & 81.20 \\
    \hline
    \end{tabular}%
  \label{tab:medical}%
\end{table}%
it can be observed from Table.~\ref{tab:medical} that the proposed GZSQ has outperformed the \textsc{ZeroQ} \cite{zeroq} and $\mathcal{N}(0,1)$ by a noticeable margin. 
\begin{table}[t]
\captionsetup{textfont={sc,footnotesize}, justification=centering, labelsep=newline}
  \centering
	    \caption{Reduction in size (in MB) of the models due to quantization. $\mathcal{M}$: Mean accuracy (INT8; QS-PT; before) on ImageNet\cite{imagenet} classification.}

    \begin{tabular}{|l|l|l|l|l|}
    \hline
    {Configs./Model} & FP32 & INT8& $\mathcal{M}$-INT8 & $\mathcal{M}$-FP32 \\
    \hline
%    \hline
    ResNet18 & 46.74 & \textbf{\textcolor{blue!95}{11.71}} & 69.27 & 69.76\\
\hline
    MobileNetV2 & 13.98 & \textbf{\textcolor{blue!95}{3.58}} & 68.50 & 71.88\\
\hline    
    ShuffleNetV2 & 9.11 & \textbf{\textcolor{blue!95}{2.34}}& 64.04 & 69.36\\
\hline    
    ResNet50 & 102.14 & \textbf{\textcolor{blue!95}{25.65}}& 75.40 & 76.15\\
\hline    
    Inception V3 & 95.32 & \textbf{\textcolor{blue!95}{23.95}}& 76.15 & 77.45\\
    \hline
    \end{tabular}%
  \label{tab:size}%
\end{table}%

\textcolor{black}{\subsection{Through the Lens of Class Activation Maps}}
Additional results are given in the provided supplementary material, wherein, Figure \ref{cams} shows the comparison among the class activation maps (CAMs) generated by utilizing the quantized models using \textsc{ZeroQ} \cite{zeroq}, and GZSQ. It can be observed that the CAMs generated by using the GZSQ's quantized model are more similar to those of original FP32 model, than those of \textsc{ZeroQ}'s. The readers are encouraged to refer Section \ref{sec:cams_details} of the supplementary material for more details. 

Table.~\ref{tab:size} shows the reduction in size (MB) of the models due to quantization.

\section{Ablation Study}
\label{ablation}
In this section, first, we deeply investigate the distilled data for various models. We then present a brief ablation study of cost functions and the effect of extra biases introduced due to BN \cite{bn} folding before distillation.

\subsection{Distilled Data}
\label{sec:ablation_dd}
Figure \ref{fig:dd_1} shows the histograms of distilled data obtained for the models that are w and w/o BN \cite{bn} layers. In Figure \ref{fig:dd_1} (Row 1), the histograms of obtained distilled data have been plotted using unit Gaussian, \textsc{ZeroQ} \cite{zeroq} and GZSQ for the ISONet \cite{isonet} model (w/o BN layers). It can be observed that the \textsc{ZeroQ} \cite{zeroq}, by its underlying backbone structure, results the distilled data similar to Gaussian but with a higher variance in the absence of BN \cite{bn} layers. This high-variance may be due to the initialization of distilled data using uniform distribution in \textsc{ZeroQ} \cite{zeroq} approach. The initialized uniform noise, in the absence of BN \cite{bn} layers, learns to shift towards a unit Gaussian within certain iterations\footnote{We encourage the readers to refer this \href{https://github.com/amirgholami/ZeroQ/blob/9715739b8731a44d064901b50260aeebbe9a10bc/classification/distill_data.py}{link} for detailed analysis.}. After finetuning, the left-out traces of uniform noise in the distilled data of the \textsc{ZeroQ} \cite{zeroq} may have caused the degradation in the accuracy for certain models, when compared to unit Gaussian and GZSQ, as shown in Tables. \ref{tab:isonets}, \ref{tab:fixup}.

Whereas, in Row 2, similar plots have been given for the MobileNetV2 \cite{mobilenetv2} model (w BN \cite{bn} layers). 
To show the effect of BN \cite{bn} folding, which has been ignored in the \textsc{ZeroQ} \cite{zeroq}, we have presented the plots for both the scenarios, when the BN \cite{bn} folding is performed before and after the data distillation. Unit Gaussian generated with different seeds is independent of the BN \cite{bn} folding. Therefore not much degradation has been observed in the case of $\mathcal{N}(0,1)$, as shown in Tables. \ref{tab:pytorch_quant_qspt}, \ref{tab:pytorch_quant_qspc}. 

However, it can be observed from Figure \ref{fig:dd_1} (Row 2) that distilled data obtained when BN \cite{bn} folding is performed before distillation is totally different when the same is performed after distillation. Especially, in the case of \textsc{ZeroQ} \cite{zeroq}, which results in a significant difference between final accuracies (see Tables. \ref{tab:pytorch_quant_qspt}, \ref{tab:pytorch_quant_qspc}). This is due to the fact that extra bias introduced because of BN \cite{bn} folding into the weights has not been considered in the \textsc{ZeroQ}'s distillation process. Consequently, it may have resulted in a poor correlation between the folded weights and the distilled data generated by looking at the BN \cite{bn} statistics prior to folding. Whereas, the same has been considered in the proposed GZSQ framework, as described in Section \ref{sec:bnf}. Thus the obtained distilled data is not only distinguishable from unit Gaussian, but also does not exhibit any severe deviation because of the BN \cite{bn} folding. Figure \ref{fig:dd_2} shows the histogram plots of the distilled data obtained for the task of pneumonia classification using ResNext101 \cite{resnext}.

%
%\begin{figure}
%\centering
%\begin{tabular}{c}
%\includegraphics[width=0.25\textwidth]{./data-plots/wbn/plot-uniform-l2/imagenet-train-sample-2.pdf}
%\end{tabular}
%\caption{Sample image from ImageNet \cite{imagenet} dataset and the corresponding manifold learning plot.}
%\end{figure}

\subsection{Effect of cost functions}
\label{cost_fn_ablation}
To show the effect of proposed $\mathcal{L}_\mathbf{Z}$ cost function (see Eqn. \ref{kd-eq}), we have presented a detailed study by performing the following baselines:
\begin{itemize}

\item \textcolor{black}{GZSQ-$\ell_1^\mu$, where $\ell_1$ norm between the means has been used as a cost function instead of proposed $\mathcal{L}_{\mathbf{Z}}$,
\begin{equation}
\| \mu_u - \mu_v \|  
\end{equation}}

\item \textcolor{black}{GZSQ-$\ell_1^\sigma$, where $\ell_1$ norm between the std-devs has been used as a cost function instead of proposed $\mathcal{L}_{\mathbf{Z}}$,
\begin{equation}
 \| \sigma_u - \sigma_v \| 
\end{equation}}

\item GZSQ-$\ell_1$, where $\ell_1$ norm has been used as a cost function instead of proposed $\mathcal{L}_{\mathbf{Z}}$,
\begin{equation}
\| \mu_u - \mu_v \|  + \| \sigma_u - \sigma_v \| 
\end{equation}

\item \textcolor{black}{GZSQ-$\ell_2^\mu$, where $\ell_2$ norm between the means has been used as a cost function instead of proposed $\mathcal{L}_{\mathbf{Z}}$,
\begin{equation}
\| \mu_u - \mu_v \|_2^2  
\end{equation}}

\item \textcolor{black}{GZSQ-$\ell_2^\sigma$, where $\ell_2$ norm between the std-devs has been used as a cost function instead of proposed $\mathcal{L}_{\mathbf{Z}}$,
\begin{equation}
 \| \sigma_u - \sigma_v \|_2^2 
\end{equation}}

\item GZSQ-$\ell_2$, where $\ell_2$ norm (similar to \textsc{ZeroQ} \cite{zeroq}) has been used as a cost function instead of proposed $\mathcal{L}_{\mathbf{Z}}$,
\begin{equation}
\| \mu_u - \mu_v \|_2^2  + \| \sigma_u - \sigma_v \|_2^2 
\end{equation}

\item GZSQ-$\mathcal{L}_{KL}^*$, where KL divergence has been used to measure the difference between two distributions instead of proposed $\mathcal{L}_{\mathbf{Z}}$. Since the estimated mean of BN \cite{bn} substitutes may have negative values (see Section \ref{sec:se}, ref: Eqn. \ref{mean_std_formula}), for which the KL divergence may be undefined. Hence, we have only utilized the std-devs to measure the difference. Therefore the adopted loss for GZSQ-$\mathcal{L}_{KL}^*$ is 
\begin{equation}
1 - 0.5* \mathbf{KL}(\sigma_u, \sigma_v),
\label{eq-kldivergence}
\end{equation}
where $\mathbf{KL}(.)$ refers to Kullback-Leibler divergence, instead of Eqn. \ref{kd-eq} \cite{kl}.
\end{itemize}
\textcolor{black}{It has been observed that for most of the cases, the parent baselines  GZSQ-$\ell_1$ and GZSQ-$\ell_2$ have shown considerable superiority over aforementioned leaf baselines GZSQ-$\ell_1^\mu$, GZSQ-$\ell_1^\sigma$, GZSQ-$\ell_2^\mu$, GZSQ-$\ell_2^\sigma$.} Further, it can be observed from Table. \ref{tab:loss_ablation} that the proposed $\mathcal{L}_{\mathbf{Z}}$ cost function has been proven to be more beneficial than the $\ell_1$ and $\ell_2$ norms. It has also been observed that the $\ell_1$ and $\ell_2$ norms are useful in restoring the structural information (\textcolor{black}{see Figure \ref{fig:cf_1}}). Whereas the proposed $\mathcal{L}_{\mathbf{Z}}$ loss restores the distributional properties of the data. To show the difference qualitatively, we present the visuals of distilled data when BN \cite{bn} folding is performed before and after the distillation phase for both \textsc{ZeroQ} \cite{zeroq} and GZSQ on ResNet18 \cite{resnet} model, in Figure \ref{fig:cf_1}. 
\begin{table}[t]
\captionsetup{textfont={sc,footnotesize}, justification=centering, labelsep=newline}
  \centering
  \caption{Mean Top-1 accuracy on various pre-trained models for the task of ImageNet \cite{imagenet} classification. Quantization configuration is INT8 with \text{QS-PC} for weights and QS-PT for activations.}
  \resizebox{0.49\textwidth}{!}{
    \begin{tabular}{|l|c|c|c|c|c|c|c|c|}
    \hline
    Model (w BN) & \textcolor{black}{GZSQ-$\ell_1^\mu$} & \textcolor{black}{GZSQ-$\ell_1^\sigma$} & \multicolumn{1}{c|}{GZSQ-$\ell_1$} & \textcolor{black}{GZSQ-$\ell_2^\mu$} & \textcolor{black}{GZSQ-$\ell_2^\sigma$} & \multicolumn{1}{c|}{GZSQ-$\ell_2$} & \multicolumn{1}{c|}{{GZSQ}} & \multicolumn{1}{c|}{FP32} \\
    \hline
%   \hline
    ResNet50 & \textcolor{black}{75.38}& \textcolor{black}{75.85} & 76.94 & \textcolor{black}{75.41} & \textcolor{black}{75.67} & 76.50  & \textcolor{blue!95}{\textbf{77.65}} & 77.72 \\
    \hline
    MobilenetV2 & \textcolor{black}{72.93} & \textcolor{black}{72.82} & 72.91 & \textcolor{black}{72.92} & \textcolor{black}{72.82} & 72.85 & \textcolor{blue!95}{\textbf{72.93}} & 73.03 \\
    \hline
    ShuffleNetV1 & \textcolor{black}{64.58} & \textcolor{black}{64.26} & 64.46 & \textcolor{black}{64.45} & \textcolor{black}{64.25} & 63.37 & \textcolor{blue!95}{\textbf{64.66}} & 65.07 \\
    \hline
    ResNet18 & \textcolor{black}{70.65} & \textcolor{black}{71.35} & 71.41 & \textcolor{black}{70.29} & \textcolor{black}{71.37} & 71.36 & \textcolor{blue!95}{\textbf{71.44}} & 71.47 \\
    \hline
    ResNet26 & \textcolor{black}{72.14} & \textcolor{black}{73.46} & 73.57 & \textcolor{black}{72.12} & \textcolor{black}{73.49} & 73.31 & \textcolor{blue!95}{\textbf{73.59}} & 73.70 \\
    \hline
    ResNet152 & \textcolor{black}{78.14} & \textcolor{black}{77.81} & 78.68 & \textcolor{black}{77.63} & \textcolor{black}{77.62} & 77.92 & \textcolor{blue!95}{\textbf{78.85}} & 80.08 \\
    \hline
    InceptionV3 & \textcolor{black}{76.83} & \textcolor{black}{76.56} & 77.16 & \textcolor{black}{74.19} & \textcolor{black}{74.25} & 75.50  & \textcolor{blue!95}{\textbf{78.79}} & 78.88 \\
    \hline
    SqueezeNextV5 & \textcolor{black}{67.81} & \textcolor{black}{67.73} & 67.31 & \textcolor{black}{67.74} & \textcolor{black}{67.03} & 65.42 & \textcolor{blue!95}{\textbf{69.32}} & 69.38 \\
    \hline
    MobilenetV1 & \textcolor{black}{72.68} & \textcolor{black}{73.05} & 73.08 & \textcolor{black}{72.79} & \textcolor{black}{73.02} & \textcolor{blue!95}{\textbf{73.10}}  & 72.99 & 73.39 \\
    \hline
    \end{tabular}}%
  \label{tab:loss_ablation}%
\end{table}%

\begin{table}[t]
\captionsetup{textfont={sc,footnotesize}, justification=centering, labelsep=newline}
\centering
\caption{Effect of $\mathcal{L}_{\mathbf{Z}}$ over KL divergence loss function in terms of ImageNet\cite{imagenet} classification. Quantization configuration is INT8 with QS-PT when BN layers are folded after distillation.}
\begin{tabular}{|l|c|c|c|}
\hline
Model (w BN) & GZSQ-$\mathcal{L}_{KL}^*$ & GZSQ & FP32\\
\hline
%\hline
ResNet18 & 63.48$\pm$0.63 & \textcolor{blue!95}{\textbf{69.09$\pm$0.05}} &69.76\\
\hline
MobileNetV2 & 64.22$\pm$0.40 & \textcolor{blue!95}{\textbf{68.55$\pm$0.17}} &71.88\\
\hline
ShuffleNetV2 & 61.00$\pm$0.14& \textbf{\textcolor{blue}{64.02$\pm$0.18}} &69.36\\
\hline
ResNet50 &74.23$\pm$0.10 & \textbf{\textcolor{blue}{75.27$\pm$0.12}} &76.15\\
\hline
InceptionV3 & 75.58$\pm$0.02 & \textbf{\textcolor{blue}{76.16$\pm$0.04}} &77.45\\
\hline
\end{tabular}
\label{tab:kl_min}
\end{table}

It can be observed that the \textsc{ZeroQ} \cite{zeroq} retains some structural information when BN \cite{bn} folding is performed after distillation phase (see Figure \ref{fig:cf_1}, \textit{row:top, col:left}). This may be due to the utilized BN \cite{bn} layer statistics. However, in this case, \textsc{ZeroQ} \cite{zeroq} in general results the poor accuracy (see Section \ref{sec:ablation_dd}). Whereas, a significant loss in structural information can be observed when folding is performed prior to distillation (see Figure \ref{fig:cf_1}, \textit{row:bottom, col:left}). Similar to \textsc{ZeroQ} \cite{zeroq}, we evaluated our baseline GZSQ-$\ell_2$ and observed that the proposed scheme is also able to retain the structural information when folding is performed after distillation phase (see Figure \ref{fig:cf_1}, \textit{row:top, col:middle}). However, in contradiction to \textsc{ZeroQ} \cite{zeroq}, the proposed baseline also retains some structural information in the distilled data when folding is performed before distillation phase (see Figure \ref{fig:cf_1}, \textit{row:bottom, col:middle}). It may be due to the incorporation of extra biases in the distillation phase, introduced due to BN \cite{bn} folding (see Section \ref{sec:bnf}), which may have carried, inherently, some information about the distribution of original training samples. 

Therefore it can be concluded that some structural information about the data can be retained even without utilizing the BN \cite{bn} statistics in the distillation phase. However, such information may not be much beneficial for some models, \textit{e.g.}, SqueezeNextV5 \cite{nndesign1}, as shown in Table. \ref{tab:loss_ablation} or \textsc{ZeroQ}'s \cite{zeroq} results for ResNet18 \cite{resnet} as shown in Table. \ref{tab:pytorch_quant_qspt}. Coming to the distributional minimization, we performed a baseline, namely, GZSQ-$\mathcal{L}_{KL}^*$, that utilizes the KL divergence instead of the proposed $\mathcal{L}_{\mathbf{Z}}$ cost function. However, it can observed from the Table. \ref{tab:kl_min} that the proposed $\mathcal{L}_{\mathbf{Z}}$ loss yields better performance than the KL divergence based cost function. It should be mentioned that the KL divergence has been utilized with half of its capability since the mean values could not be incorporated in Eqn. \ref{eq-kldivergence}. Though, it may be taken as a slight drawback of GZSQ. But, nothing for certain can be concluded about the end performance if KL divergence would have been utilized fully against the proposed $\mathcal{L}_{\mathbf{Z}}$.

It can also be observed from Figure \ref{fig:cf_1} (\textit{row:top/bottom, col:right}) that the stills of distilled data, when utilizing GZSQ, may not comprise of much structural information. However, it retains rich distributional properties that helps in achieving the near FP32 accuracy post quantization, as shown in Table. \ref{tab:loss_ablation}. When comparing with unit Gaussian samples, the inclusion of pre-trained weights in the distillation process of GZSQ may have acted as the cherry on the top. 

\begin{table}[t]
\captionsetup{textfont={sc,footnotesize}, justification=centering, labelsep=newline}
  \centering
    \caption{Mean Top-1 accuracy for the ImageNet \cite{imagenet} classification when BN folding is done before distillation. Quant. config.: INT8, \text{QS-PC}/\text{QS-PT} for weights, and \text{QS-PT} for activations.}
      
  \resizebox{0.48\textwidth}{!}{
    \begin{tabular}{|l|c|c|c|c|c|}
\hline
    \multirow{3}[0]{*}{ Model (\text{w} BN) } & \multicolumn{4}{c|}{ \text{GZSQ}}      & \multirow{3}[0]{*}{ FP32} \\
    
    \cline{2-5}

    \multicolumn{1}{|c|}{} & \multicolumn{2}{c|}{ QS-PC} & \multicolumn{2}{c|}{ QS-PT} &  \\

    \cline{2-5}
    \multicolumn{1}{|c|}{} & { w Bias} & { w/o Bias} & { w Bias} & { w/o Bias} &  \\

    \hline
%    \hline
    ResNet18 & \textcolor{blue!95}{{\textbf{69.47}}} & 44.22 & \textcolor{blue!95}{{\textbf{69.27}}} &   44.61    & 69.76 \\
    \hline
    MobileNetV2 & 69.38 & \textcolor{blue!95}{{\textbf{69.68}}} & 68.50 & \textcolor{blue!95}{{\textbf{68.66}}} & 71.88 \\
    \hline
    ShuffleNetV2 & 67.72 & \textcolor{blue!95}{{\textbf{67.78}}} & \textcolor{blue!95}{\textbf{64.04}} &  63.97   & 69.36 \\
    \hline
    ResNet50 & \textcolor{blue!95}{{\textbf{75.83}}} & 75.63 & \textcolor{blue!95}{{\textbf{75.40}}} &    75.26   & 76.15 \\
 \hline
    InceptionV3 & \textcolor{blue!95}{{\textbf{77.09}}} & 77.08 & 76.15 &  \textcolor{blue!95}{{\textbf{76.18}}}     & 77.45 \\
%    \hline
\hline
    \end{tabular}}%
%\vspace*{-0.1cm}

  \label{tab:bias_vs_no_bias}%
\end{table}%

\subsection{Effect of BN folding}
\label{sec:effect_of_bnf}
In Table~\ref{tab:bias_vs_no_bias}, we have shown the results to demonstrate the effect of extra biases, introduced due to BN\cite{bn} folding, when considered or ignored during our distillation process. It can be observed that for most of models, especially in the case of ResNet18 \cite{resnet}, incorporating extra biases in the distillation phase (see Section \ref{sec:bnf}) yields better results than ignoring the same in the distillation phase, irrespective of the quantization configuration. The incorporated biases may have reduced the correlation shift between the folded weights and distilled data, resulting in a near-accurate estimation of activation ranges to calculate quantization parameters.  

\section{Conclusions}
\label{conclusion}
In this work, we have presented a novel \textit{generalized zero-shot quantization} (\textsc{GZSQ}\xspace) framework for the post-training quantization of deep CNNs, that leverages only the pre-trained weights of the model. The proposed zero-shot approach does not rely on the original unlabelled data or the learned BN layer parameters to infer the activation ranges. Our proposed scheme is built upon the data distillation approach. And, we argued that {Z-score} based loss could be more effective than $\ell_1$ and $\ell_2$ norms or KL divergence, during distillation. 

We have benchmarked the models (w \& w/o BN layers) for the tasks of classification and object detection. We have also presented the quantization results for pneumonia classification in chest X-rays. We have shown that our method performed well, even when BN layer is folded or absent. Further, our approach is more generalized, outperforming the best-published works across multiple quantization frameworks. We have also presented a detailed ablation study, demonstrating the effect of various cost functions and BN folding. 

The presented work is the first attempt towards the post-training zero-shot quantization of the futuristic unnormalized deep CNNs. And, in the subsequent time, we would like to extend our approach to mixed precision quantization and other problem domains. For \textit{e.g.}, multi-modal classification, image de-noising \& reconstruction, a glimpse of whose foundation has been given in the Figure \ref{deraining} of the supplemental file.

\section*{acknowledgement}
The authors would like to thank the anonymous Reviewers
and Editors for their insightful observations, comments and
suggestions, which helped to improve the quality of this work.

{\footnotesize
\bibliographystyle{IEEEtran}
\bibliography{refs}
}

\newpage
\begin{figure*}[t]
\centering
\setlength{\tabcolsep}{1pt}
\resizebox{\textwidth}{!}{
    \begin{tabular}{cccccccc}

\textcolor{black}{PC}& &  \multicolumn{2}{c}{\textcolor{black}{\texttt{\textbf{Packet: 5.78}}}} & \multicolumn{2}{c}{\textcolor{black}{\texttt{Schooner: 5.55}}} & \multicolumn{2}{c}{\textcolor{black}{\texttt{Schooner: 5.79}}}\\
  
\rotatebox{90}{\textcolor{black}{\texttt{Schooner}}}&\includegraphics[width=2.4cm, height=1.8cm]{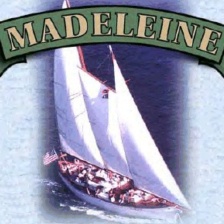} 
&  
\includegraphics[width=2.4cm, height=1.8cm]{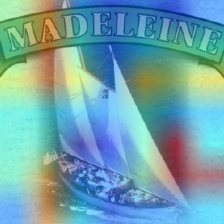} 
& 
\includegraphics[width=2.4cm, height=1.8cm]{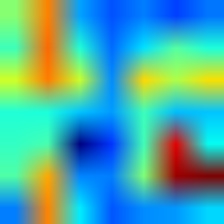}  
& 
\includegraphics[width=2.4cm, height=1.8cm]{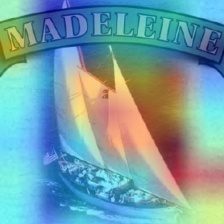} 
& 
\includegraphics[width=2.4cm, height=1.8cm]{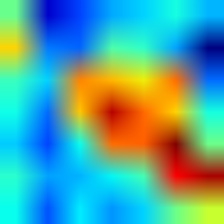}    
& 
\includegraphics[width=2.4cm, height=1.8cm]{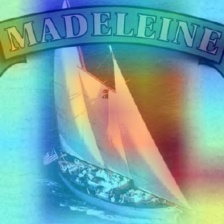} 
&
\includegraphics[width=2.4cm, height=1.8cm]{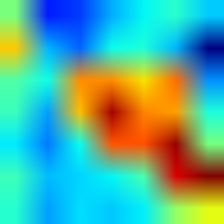} \\

& &  \multicolumn{2}{c}{\textcolor{black}{\texttt{Zebra: 9.31}}} & \multicolumn{2}{c}{\textcolor{black}{\texttt{Zebra: 9.46}}} & \multicolumn{2}{c}{\textcolor{black}{\texttt{Zebra: 9.62}}}\\

\rotatebox{90}{\textcolor{black}{~~\texttt{Zebra}}}&\includegraphics[width=2.4cm, height=1.8cm]{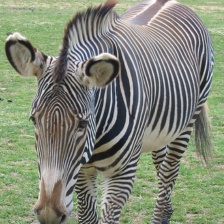} 
&  
\includegraphics[width=2.4cm, height=1.8cm]{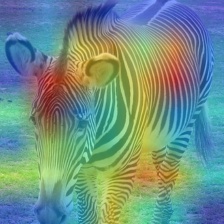} 
& 
\includegraphics[width=2.4cm, height=1.8cm]{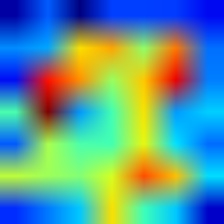}  
& 
\includegraphics[width=2.4cm, height=1.8cm]{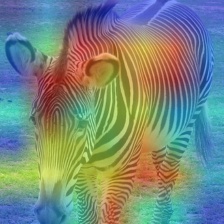} 
& 
\includegraphics[width=2.4cm, height=1.8cm]{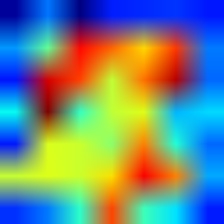}    
& 
\includegraphics[width=2.4cm, height=1.8cm]{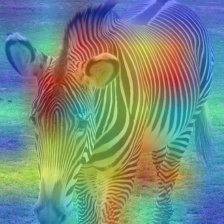} 
&
\includegraphics[width=2.4cm, height=1.8cm]{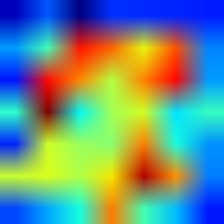} \\

& &  \multicolumn{2}{c}{\textcolor{black}{\texttt{\textbf{Warplane: 9.06}}}} & \multicolumn{2}{c}{\textcolor{black}{\texttt{\textbf{Warplane: 7.83}}}} & \multicolumn{2}{c}{\textcolor{black}{\texttt{\textbf{Warplane: 8.52}}}}\\

\rotatebox{90}{\textcolor{black}{\texttt{Aircraft}}}&\includegraphics[width=2.4cm, height=1.8cm]{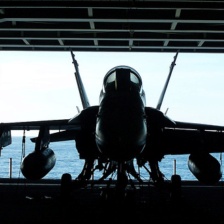} 
&  
\includegraphics[width=2.4cm, height=1.8cm]{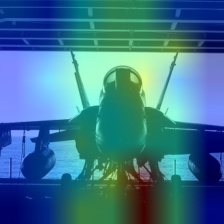} 
& 
\includegraphics[width=2.4cm, height=1.8cm]{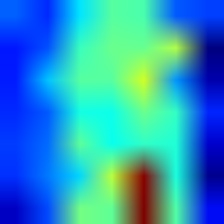}  
& 
\includegraphics[width=2.4cm, height=1.8cm]{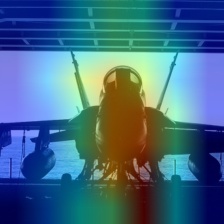} 
& 
\includegraphics[width=2.4cm, height=1.8cm]{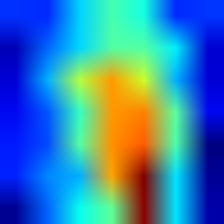}    
& 
\includegraphics[width=2.4cm, height=1.8cm]{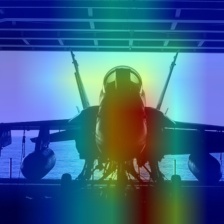} 
&
\includegraphics[width=2.4cm, height=1.8cm]{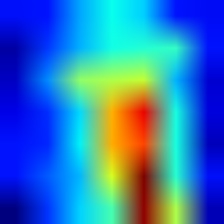} \\

& &  \multicolumn{2}{c}{\textcolor{black}{\texttt{Monkey: 9.55}}} & \multicolumn{2}{c}{\textcolor{black}{\texttt{Monkey: 9.82}}} & \multicolumn{2}{c}{\textcolor{black}{\texttt{Monkey: 9.67}}}\\

\rotatebox{90}{\textcolor{black}{~~\texttt{Monkey}}}&\includegraphics[width=2.4cm, height=1.8cm]{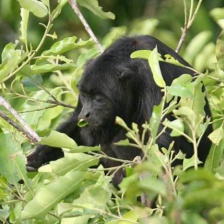} 
&  
\includegraphics[width=2.4cm, height=1.8cm]{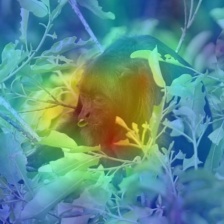} 
& 
\includegraphics[width=2.4cm, height=1.8cm]{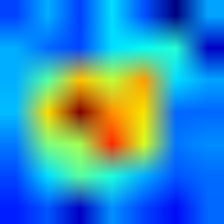}  
& 
\includegraphics[width=2.4cm, height=1.8cm]{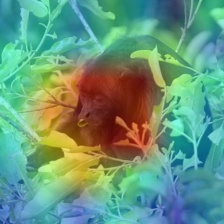} 
& 
\includegraphics[width=2.4cm, height=1.8cm]{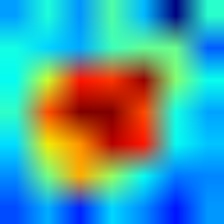}    
& 
\includegraphics[width=2.4cm, height=1.8cm]{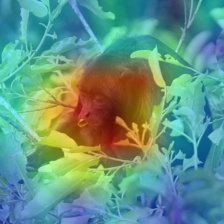} 
&
\includegraphics[width=2.4cm, height=1.8cm]{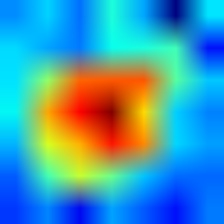} \\

& &  \multicolumn{2}{c}{\textcolor{black}{\texttt{Sandbar: 9.76}}} & \multicolumn{2}{c}{\textcolor{black}{\texttt{Sandbar: 9.87}}} & \multicolumn{2}{c}{\textcolor{black}{\texttt{Sandbar: 9.34}}}\\

\rotatebox{90}{\textcolor{black}{~~\texttt{Sandbar}}}&\includegraphics[width=2.4cm, height=1.8cm]{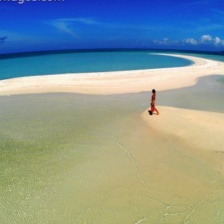} 
&  
\includegraphics[width=2.4cm, height=1.8cm]{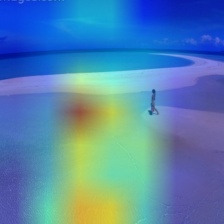} 
& 
\includegraphics[width=2.4cm, height=1.8cm]{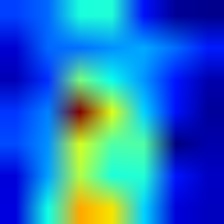}  
& 
\includegraphics[width=2.4cm, height=1.8cm]{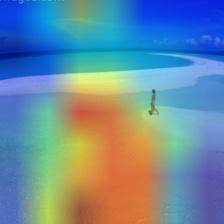} 
& 
\includegraphics[width=2.4cm, height=1.8cm]{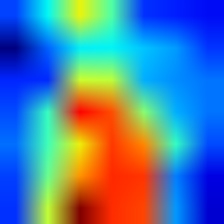}    
& 
\includegraphics[width=2.4cm, height=1.8cm]{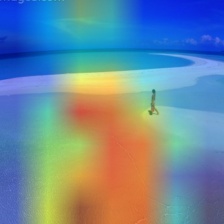} 
&
\includegraphics[width=2.4cm, height=1.8cm]{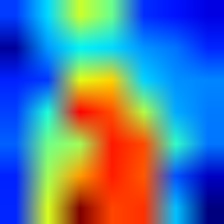} \\

& &  \multicolumn{2}{c}{\textcolor{black}{\texttt{Candle: 8.25}}} & \multicolumn{2}{c}{\textcolor{black}{\texttt{Candle: 8.80}}} & \multicolumn{2}{c}{\textcolor{black}{\texttt{Candle: 8.50}}}\\

\rotatebox{90}{\textcolor{black}{~~\texttt{Candle}}}&\includegraphics[width=2.4cm, height=1.8cm]{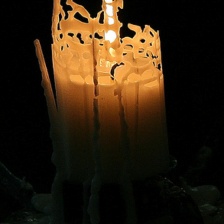} 
&  
\includegraphics[width=2.4cm, height=1.8cm]{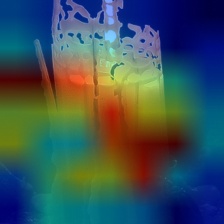} 
& 
\includegraphics[width=2.4cm, height=1.8cm]{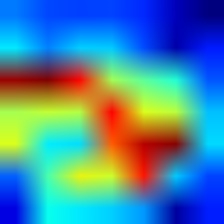}  
& 
\includegraphics[width=2.4cm, height=1.8cm]{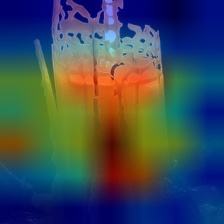} 
& 
\includegraphics[width=2.4cm, height=1.8cm]{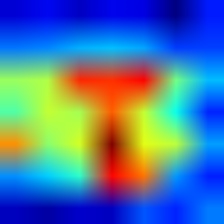}    
& 
\includegraphics[width=2.4cm, height=1.8cm]{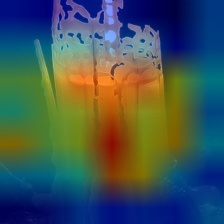} 
&
\includegraphics[width=2.4cm, height=1.8cm]{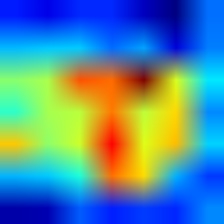} \\

& &  \multicolumn{2}{c}{\textcolor{black}{\texttt{Model T: 12.02}}} & \multicolumn{2}{c}{\textcolor{black}{\texttt{Model T: 11.78}}} & \multicolumn{2}{c}{\textcolor{black}{\texttt{Model T: 11.90}}}\\

\rotatebox{90}{\textcolor{black}{~~\texttt{Model T}}}&\includegraphics[width=2.4cm, height=1.8cm]{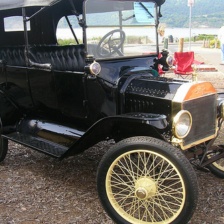} 
&  
\includegraphics[width=2.4cm, height=1.8cm]{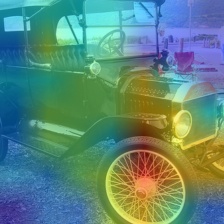} 
& 
\includegraphics[width=2.4cm, height=1.8cm]{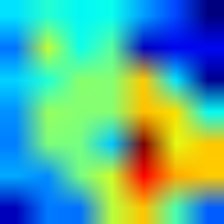}  
&
\includegraphics[width=2.4cm, height=1.8cm]{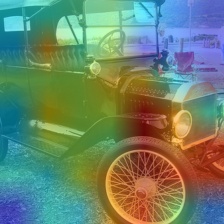} 
& 
\includegraphics[width=2.4cm, height=1.8cm]{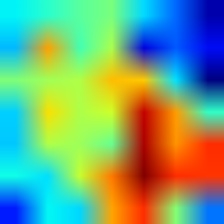}    
& 
\includegraphics[width=2.4cm, height=1.8cm]{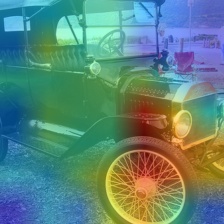} 
&
\includegraphics[width=2.4cm, height=1.8cm]{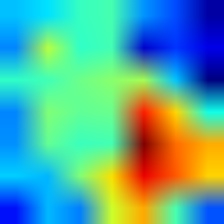} \\

& &  \multicolumn{2}{c}{\textcolor{black}{\texttt{\textbf{Submarine: 8.38}}}} & \multicolumn{2}{c}{\textcolor{black}{\texttt{Shp. Cart: 8.24}}} & \multicolumn{2}{c}{\textcolor{black}{\texttt{Shp. Cart: 8.22}}}\\

\rotatebox{90}{\textcolor{black}{\texttt{Shp. Cart}}}&\includegraphics[width=2.4cm, height=1.8cm]{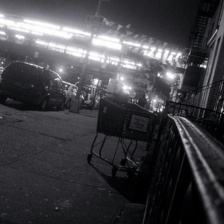} 
&  
\includegraphics[width=2.4cm, height=1.8cm]{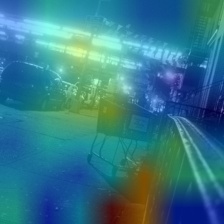} 
& 
\includegraphics[width=2.4cm, height=1.8cm]{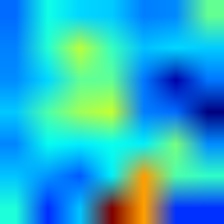}  
& 
\includegraphics[width=2.4cm, height=1.8cm]{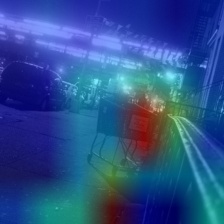} 
& 
\includegraphics[width=2.4cm, height=1.8cm]{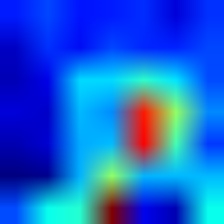}    
& 
\includegraphics[width=2.4cm, height=1.8cm]{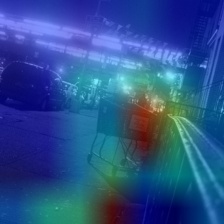} 
&
\includegraphics[width=2.4cm, height=1.8cm]{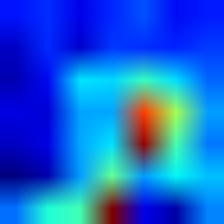} \\

\textcolor{black}{TC}&\texttt{Input} & \multicolumn{2}{c}{\textsc{\texttt{ZeroQ}} \cite{zeroq}} & \multicolumn{2}{c}{\texttt{GZSQ}} & \multicolumn{2}{c}{\texttt{FP32}}\\    

\multicolumn{8}{c}{\includegraphics[width=\textwidth, height=0.3cm]{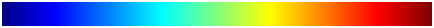}}\\
          
    \end{tabular}}%
%  \label{tab:addlabel}%
\caption{Sample stills of class activation maps on the ImageNet \cite{imagenet} dataset using the quantized models from \textsc{ZeroQ} \cite{zeroq},  and GZSQ. FP32 denotes the activation maps generated by using the original floating point model. We have utilized the SqueezeNextV5 \cite{nndesign1} for this small study.}
\label{cams}
\end{figure*}%
\clearpage
\section{\textcolor{black}{Through the lens of Class Activation Maps (cont.)}}
\label{sec:cams_details}
\textcolor{black}{On an abstract level, class activation maps depict the regions in the input image, which have been addressed by the deep models with more focus during classification or any other similar high-level vision task. Figure \ref{cams} shows the class activation maps generated by using the quantized models for the task of Imagenet \cite{imagenet} classification. For this, we have utilized the SqueezeNextV5 \cite{nndesign1} model as FP32 baseline. This brief study aims to demonstrate the effect of quantization on the model's focus through the lens of activation maps. In Figure \ref{cams}, PC denotes the predicted class by the quantized model, whereas TC refers to the true class of the input image. It can be observed that across a variety of classes, activation maps generated by the quantized model using GZSQ are more similar to FP32 maps when compared to \textsc{ZeroQ} \cite{zeroq} maps. Further, the scores generated by the GZSQ quantized model are more aligned towards those of the FP32 against \textsc{ZeroQ} \cite{zeroq}. The warmer colors in the activation map denote the regions with more focus, whereas the lighter colors show the regions with less emphasis. Therefore, in terms of priority, it can be concluded that the quantized model using the GZSQ scheme performs more similarly to the original FP32 than the \textsc{ZeroQ}'s \cite{zeroq} quantized model. }

\section{\textcolor{black}{Single Image De-Raining using Quantized Models}}

\begin{table}[h]
\captionsetup{textfont={sc,footnotesize}, justification=centering, labelsep=newline}
\caption{\textcolor{black}{Quantitative results on de-rained images generated by the quantized/FP32 models using \texttt{Real Internet} dataset \cite{Wang_2019_CVPR} in terms of non-reference image quality metrics with \texttt{L:} lower is better. Best results are highlighted with bold fonts.}}
\label{tab:rain}
\begin{center}
\resizebox{\linewidth}{!}{
\begin{tabular}{|l|c|c|c|c|c|c|}
\hline
\textcolor{black}{Metrics} & \textcolor{black}{Behaviour} & \textcolor{black}{Input} & \textcolor{black}{$\mathcal{N}(0,1)$} & \textsc{\textcolor{black}{ZeroQ}} \cite{zeroq} & \textcolor{black}{GZSQ} & \textcolor{black}{FP32}\\
\hline
\textcolor{black}{NIQE} & \textcolor{black}{\texttt{L}} & \textcolor{black}{4.7918} & \textcolor{black}{3.8296} & \textcolor{black}{3.8219} & \textcolor{black}{\textbf{3.8162}} & \textcolor{black}{3.6905} \\
\hline

\textcolor{black}{BRISQUE} & \textcolor{black}{\texttt{L}} & \textcolor{black}{32.0835} & \textcolor{black}{14.8261} & \textcolor{black}{15.1901} & \textcolor{black}{\textbf{14.7678}} & \textcolor{black}{13.3711} \\
\hline

\textcolor{black}{TV Error} & \textcolor{black}{\texttt{L}} & \textcolor{black}{7.02$\times 10^6$} & \textcolor{black}{6.48$\times 10^6$} & \textcolor{black}{\textbf{6.42$\times 10^6$}} & \textcolor{black}{6.50$\times 10^6$} & \textcolor{black}{5.26$\times 10^6$} \\
\hline

\end{tabular}}
\end{center}
\end{table}
\begin{figure}[h]
\centering
\setlength{\tabcolsep}{2pt}
\resizebox{\linewidth}{!}{
\begin{tabular}{ccc}
{\scriptsize \textit{NIQE: 3.4353}} & {\scriptsize \textit{NIQE: 3.2307}} & {\scriptsize \textit{NIQE: 2.9743}} \\ 
\includegraphics[width=2.4cm, height=1.8cm]{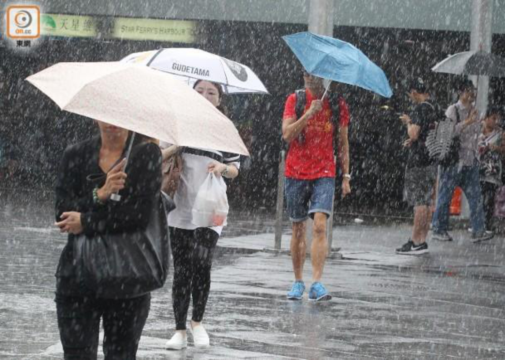}&
\includegraphics[width=2.4cm, height=1.8cm]{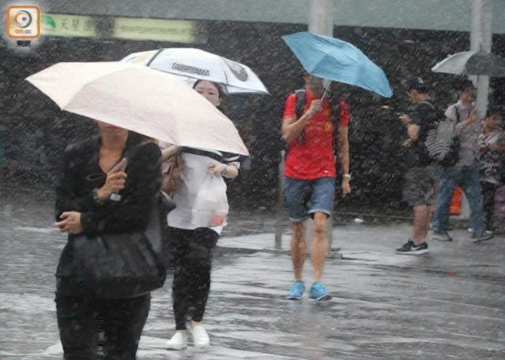}&
\includegraphics[width=2.4cm, height=1.8cm]{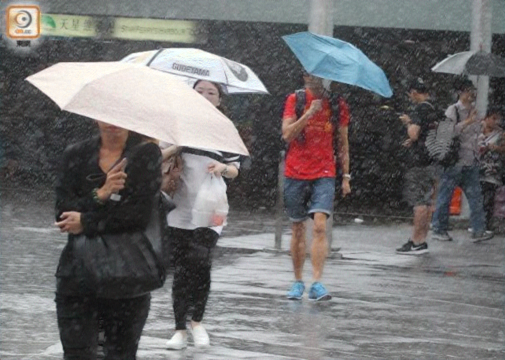}\\
{\scriptsize \textit{NIQE: 5.4935}} & {\scriptsize \textit{NIQE: 3.7871}} & {\scriptsize \textit{NIQE: 3.6622}}\\
\includegraphics[width=2.4cm, height=1.8cm]{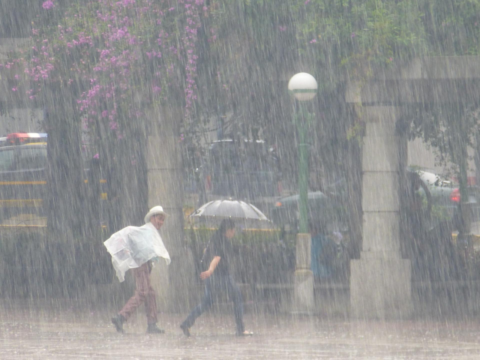}&
\includegraphics[width=2.4cm, height=1.8cm]{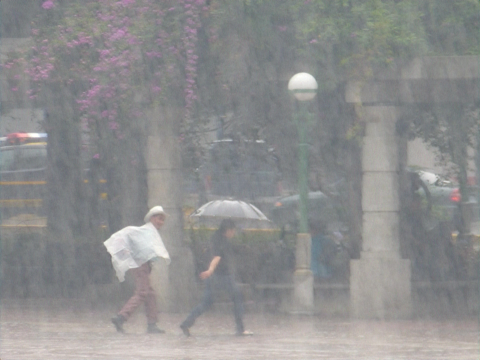}&
\includegraphics[width=2.4cm, height=1.8cm]{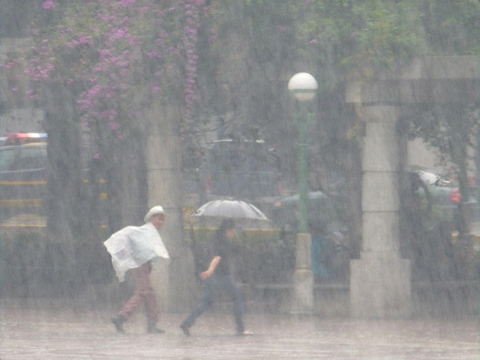}\\

{\scriptsize \textit{NIQE: 3.5157}} & {\scriptsize \textit{NIQE: 2.7773}} & {\scriptsize \textit{NIQE: 2.5181}} \\ 
\includegraphics[width=2.4cm, height=1.8cm]{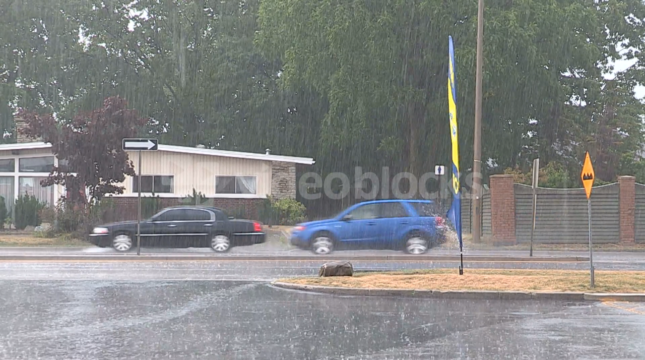}&
\includegraphics[width=2.4cm, height=1.8cm]{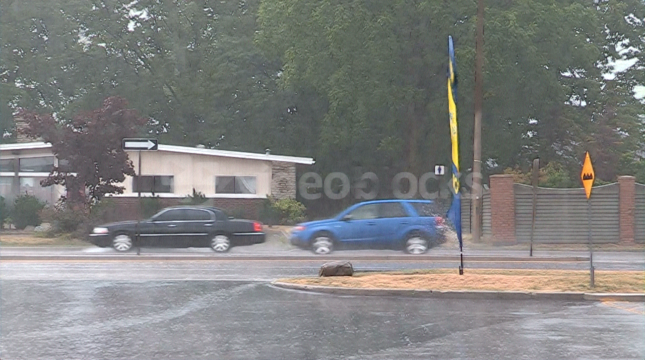}&
\includegraphics[width=2.4cm, height=1.8cm]{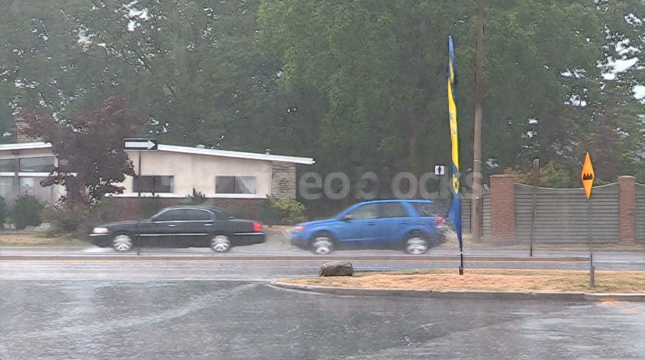}\\
{\scriptsize \textit{NIQE: 3.6045}} & {\scriptsize \textit{NIQE: 3.1190}} & {\scriptsize \textit{NIQE: 2.9183}}\\
\includegraphics[width=2.4cm, height=1.8cm]{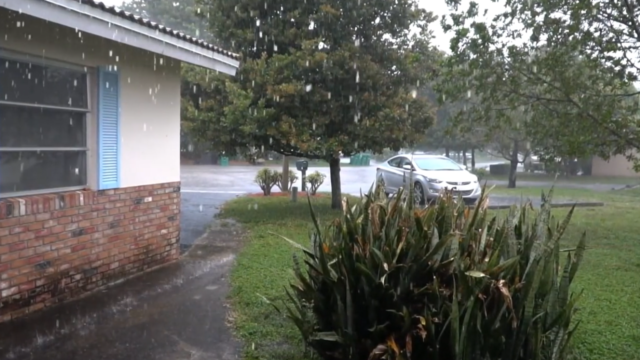}&
\includegraphics[width=2.4cm, height=1.8cm]{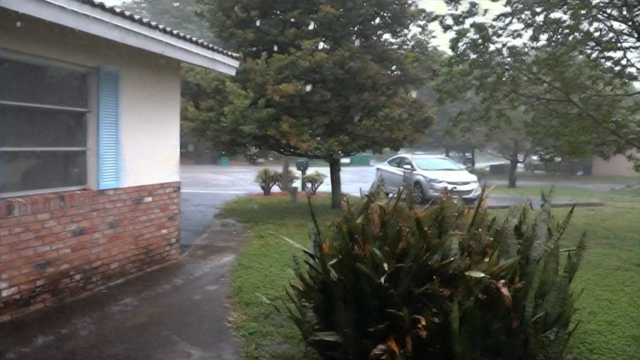}&
\includegraphics[width=2.4cm, height=1.8cm]{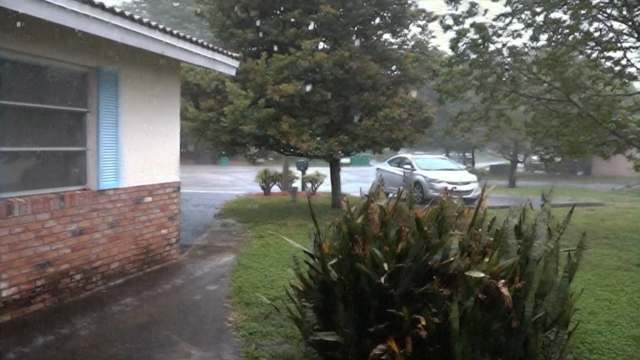}\\

 & {\scriptsize \textsc{ZeroQ} \cite{zeroq}} & {\scriptsize GZSQ} \\  
% & {\scriptsize \textsc{ZeroQ} \cite{zeroq}} & {\scriptsize GZSQ}\\

{\scriptsize \texttt{Rainy}} & \multicolumn{2}{c}{{\scriptsize \texttt{De-rained images}}} \\
%{\scriptsize Rainy} & \multicolumn{2}{c}{{\scriptsize De-rained images}}\\

\end{tabular}}
\caption{Sample results on single image rain-streak removal by using the quantized models (INT8) from \textsc{ZeroQ} \cite{zeroq} and GZSQ in terms of Naturalness Image Quality Evaluator (NIQE; \textit{smaller is better}). The original baseline FP32 model is DRN \cite{drn}.}
\label{deraining}
\end{figure}
\textcolor{black}{
To show the robustness of the GZSQ across different domains, we have presented a brief ablation study on single image de-raining using quantized models. For this, we have adopted  the pre-trained DRN \cite{drn} model as FP32 baseline and perform INT8 quantization using $\mathcal{N}(0,1)$, \textsc{ZeroQ} \cite{zeroq} and GZSQ. For quantitative analysis, we have utilized \texttt{Real Internet} \cite{Wang_2019_CVPR} dataset that consists of 146 real-world rainy images collected from the internet. The selected real-world rainy images do not have ground truth; hence for a fair comparison, we have utilized the reference-less image quality metrics, namely, Naturalness Image Quality Evaluator (NIQE) \cite{niqe}, Blind/Referenceless Image Spatial Quality Evaluator (BRISQUE) \cite{brisque}, and Total Variation Error (TV). The average spatial dimension of the images in the incorporated dataset is 371 px in height and 568 px in width. In general, image de-raining models trained on synthetic datasets tend to learn a different distribution of rain-streaks, unlike in real-world images, which is why most deep learning-based methods for image de-raining do not generalize well. However, in this paper, we restrict our analysis to the effect of quantization only. It can be observed from Table \ref{tab:rain} that the results obtained by using GZSQ's quantized model have surpassed \textsc{ZeroQ} \cite{zeroq} in terms of NIQE \cite{niqe} and BRISQUE \cite{brisque}. We have also presented the qualitative results in Figure \ref{deraining} in terms of NIQE \cite{niqe} against \textsc{ZeroQ} \cite{zeroq}.
}
%\clearpage

\end{document}